\documentclass[10pt]{article}
\usepackage{amsfonts, amssymb, amsmath} 
\usepackage{enumitem} 
\usepackage[left=1.00in,right=1.00in,top=1.00in,bottom=1.00in]{geometry}
\usepackage{setspace} 
\usepackage{graphicx} 
\usepackage{subfig} 
\usepackage{float} 
\usepackage{url} 
\usepackage{indentfirst} 
\usepackage[normalem]{ulem} 
\usepackage{cite} 
\usepackage{tabularray} 
\UseTblrLibrary{booktabs} 
\usepackage[title]{appendix} 
\usepackage{etoolbox, apptools} 
\AtBeginEnvironment{appendices}{\appendixtrue}

\usepackage{titlesec}

\titleformat{\section}[block]  
  {\fontsize{11}{13}\bfseries\filcenter} 
  {\IfAppendix{\appendixname~}{\relax}\thesection\IfAppendix{: }{}} 
  {0.5em} 
  {} 
  [] 

\titleformat{\subsection}  
  {\fontsize{10}{12}\bfseries} 
  {\thesubsection} 
  {0.5em} 
  {} 
  [] 

\DeclareMathOperator*{\argmin}{argmin}
\DeclareMathOperator*{\swish}{swish}
\begin{document}

\begin{center}
{\fontsize{17}{0}\selectfont \textbf{Epistemic Modeling Uncertainty of Rapid Neural Network Ensembles for Adaptive Learning}}


\vspace{17 pt}

Atticus Beachy\footnote{PhD Candidate, Department of Mechanical and Materials Engineering. Corresponding Author, \textit{Email address:} beachy.10@wright.edu} and Harok Bae\footnote{Associate Professor, Department of Mechanical and Materials Engineering.}

\textit{Wright State University, Dayton, OH, 45435, USA} 

\quad

Jose A. Camberos\footnote{Associate Professor, Department of Aeronautics and Astronautics.} and Ramana V. Grandhi\footnote{Professor, Department of Aeronautics and Astronautics.}

\textit{Air Force Institute of Technology, Wright-Patterson AFB, OH, 45433, USA} 

\quad

\end{center}

\leftskip0.5in\relax
\rightskip0.5in\relax
\textbf{Emulator embedded neural networks, which are a type of physics informed neural network, leverage multi-fidelity data sources for efficient design exploration of aerospace engineering systems. Multiple realizations of the neural network models are trained with different random initializations. The ensemble of model realizations is used to assess epistemic modeling uncertainty caused due to lack of training samples. This uncertainty estimation is crucial information for successful goal-oriented adaptive learning in an aerospace system design exploration. However, the costs of training the ensemble models often become prohibitive and pose a computational challenge, especially when the models are not trained in parallel during adaptive learning. In this work, a new type of emulator embedded neural network is presented using the rapid neural network paradigm. Unlike the conventional neural network training that optimizes the weights and biases of all the network layers by using gradient-based backpropagation, rapid neural network training adjusts only the last layer connection weights by applying a linear regression technique. It is found that the proposed emulator embedded neural network trains near-instantaneously, typically without loss of prediction accuracy. The proposed method is demonstrated on multiple analytical examples, as well as an aerospace flight parameter study of a generic hypersonic vehicle.}

\leftskip0in\relax
\rightskip0in\relax

\vspace{2.0em}

\noindent \textit{Keywords:}

\noindent Machine Learning, ~ Neural Network, ~ Multifidelity, ~ Active Learning, ~ Aircraft Design

\vspace{1.0em}

\section{Introduction}

Design exploration for high-performance aircraft requires computational modeling of multiple unconventional design configurations. Models must capture aerodynamics, structure, and propulsion, as well as interactions between these disciplines. A common design exploration technique is to sample the expensive physics-based models in a design of experiments and then use the sample data to train an inexpensive metamodel. Conventional metamodels include regression models such as ridge regression \cite{hoerl_ridge_1970}, Lasso \cite{tibshirani_regression_1996}, Polynomial Chaos Expansion \cite{rumpfkeil_multi-fidelity_2019}, Gaussian process regression (GPR) or kriging \cite{jones_efficient_1998, moore_value-based_2014, bae_nondeterministic_2019}, and neural networks \cite{alhazmi_training_2021}. However, many simulation evaluations are needed for the design of experiments because of the large number of independent parameters for each design and the complex responses resulting from interactions across multiple disciplines. Because high-fidelity simulations are expensive, the total computational costs can easily become computationally intractable.

Computational cost reduction is often achieved using Multi-Fidelity Methods (MFM) and Active Learning (AL). MFMs work by supplementing High-Fidelity (HF) simulations with less accurate but inexpensive Low-Fidelity (LF) simulations. AL involves intelligent generation of training data in an iterative process: rebuilding the metamodel after each HF sample is added, and then using the metamodel to determine the best HF sample to add next. 

When performing Multi-Fidelity (MF) modeling \cite{peherstorfer_survey_2018, fischer_bayesian-enhanced_2018}, the usual strategy is to generate many affordable Low-Fidelity (LF) samples to capture the design space and correct them using a small number of expensive High-Fidelity (HF) samples. MF modeling is a more cost-effective way of training an accurate surrogate model than using a single fidelity, which will suffer from sparseness with only HF samples and inaccuracy with only LF samples. One method for generating LF data is Model Order Reduction (MOR). Nachar et al. \cite{nachar_multi-fidelity_2020} used this technique to generate LF data for a multi-fidelity kriging model. Reduced order models \cite{shah_finite_2022, antil_novel_2021, carlberg_galerkin_2017} are constructed by approximating the low-dimensional manifold on which the solutions lie. The manifold can be approximated linearly using Proper Orthogonal Decomposition, or nonlinearly using various methods such as an autoencoder neural network \cite{boncoraglio_active_2021}. This approximation enables vastly decreasing the model degrees of freedom, which in turn reduces the computational costs of a simulation. Typically, reduced order models are used to give final predictions instead of being used as LF models in a multi-fidelity surrogate. However, this limits the aggressiveness with which the model can be reduced without introducing unacceptable errors into the final predictions. Even after very aggressive reduction, reduced order models can still provide valuable information about trends when used as LF models. LF models can also be constructed by coarsening the mesh, simplifying the physics, or utilizing historical data from a similar problem. 

When performing AL, location-specific epistemic uncertainty information is critical for determining where to add additional samples. Kriging is popular in large part because it returns this uncertainty. Cokriging \cite{le_gratiet_recursive_2014, ranftl_bayesian_2019} is a popular type of MFM which extends kriging to multiple fidelities. It typically performs well but has difficulties if the LF function is not well correlated with the HF function. It also cannot incorporate more than a single LF function unless they fall into a strict hierarchy known beforehand. As an alternative to cokriging, a localized-Galerkin kriging approach has been developed which can combine multiple nonhierarchical LF functions and enable adaptive learning \cite{bae_multifidelity_2020, beachy_expected_2020}. However, kriging and kriging-based methods have fundamental numerical limitations. Fitting a kriging model requires optimization of the hyperparameters $\theta$, where a different $\theta$ parameter exists for each dimension. Additionally, each evaluation of the loss function requires inverting the covariance matrix, an operation with computational complexity on the order of $O(d^3)$, where $d$ is the number of dimensions. This makes kriging-based methods poorly suited for modeling functions above a couple of dozen dimensions, especially if the number of training samples is high. 

\begin{figure}[h]
\centering
\includegraphics[width = 0.45\textwidth]{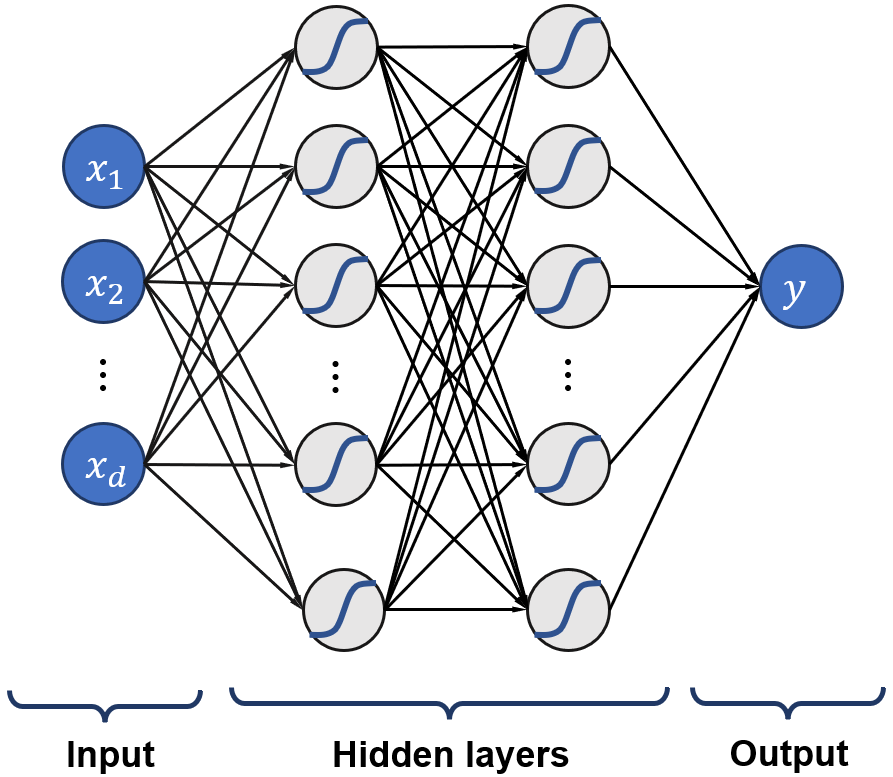}
\caption{Illustration of a fully connected multi-layer NN. A single hidden layer with sufficiently many neurons will enable a NN to model any continuous function.}
\label{fig:nn}
\end{figure}

In this work, we use Neural Networks (NNs) to avoid the limitations of kriging. A common NN architecture, feed forward fully-connected, is shown in Fig. \ref{fig:nn}. A neural network is a structure composed of layers of neurons with weighted connections to the neurons of other layers. Each neuron takes as input the weighted sum of neurons in the previous layer, plus the neuron's bias term. This input is then transformed by the neuron's activation function, and output in turn to the neurons of the next layer. 

Training a NN involves adjusting the many weights and biases to output accurate predictions on training data. Typically, weight optimization is performed using a gradient descent algorithm such as Adam, with the gradients calculated via backpropagation.

The more weights and biases a NN has, the higher the degree of flexibility. This allows NNs to function as universal approximators. Specifically, the Universal Approximation Theorem (UAT) states that a feed-forward neural network with a single hidden layer can approximate any continuous function to any degree of accuracy over a bounded region if sufficiently many neurons with nonpolynomial activation functions are included in the hidden layer \cite{cybenko_approximation_1989, hornik_approximation_1991, leshno_multilayer_1993}. However, the UAT does not put upper bounds on the error of a neural network with a limited number of neurons. Therefore, in practice, the number of layers and neurons in each layer is typically selected based on experience or trial and error. 

Unfortunately, using conventional NNs for aerospace engineering design studies faces several practical drawbacks. In aerospace applications, physical experiments or high-fidelity simulations are typically expensive, meaning HF data will be sparse. Therefore, design studies often require extrapolating beyond the bounds of the HF data, which interpolators such as NNs are poorly suited to. 

One approach for mitigating these practical problems is to use Physics Informed Neural Networks (PINNs) \cite{peng_multiscale_2021, raissi_physics-informed_2019}. These combine physics information with a NN model. Typically, the physics information takes the form of differential equations, such as governing equations and boundary conditions. PINNs have various structures, such as NNs modifying parameters of the differential equations to improve accuracy, or NNs adding corrections and detail on top of a simplified model that does not capture the full physics. The physics information constrains the model, alleviating overfitting and increasing the accuracy of extrapolations beyond the HF data points. 

PINNs can also use physics models other than differential equations, such as first-principle models, data-driven models, and expert knowledge models. PINNs can also include multiple fidelities of data, for instance, by training a NN on LF training data and using the output as an input to the PINN trained on HF data \cite{meng_composite_2020, liu_multi-fidelity_2019, aydin_general_2019}. 

Recently, the authors proposed the Emulator Embedded Neural Network (E2NN) \cite{beachy_emulator_2021, beachy_emulator_2021-1} a generic framework for combining any mix of physics-based models, experimental data, and any other information sources. Instead of using an LF model as an input to the main NN, E2NN embeds emulator models into its hidden layer neurons intrusively. A neuron with an LF model embedded into it is called an emulator. The E2NN is trained to transform and mingle the information from multiple emulators within the NN by finding the optimal connection weights and biases. This has an effect like that of a standard PINN architecture, reducing overfitting and enabling accurate extrapolations beyond sparse HF data points. E2NN performed well in the tests of the reference papers, handling non-stationary function behaviors, capturing a high dimensional Rosenbrock function with low-quality LF models, and successfully modeling stress in a Generic Hypersonic Vehicle (GHV) engineering example. 

To enable AL, it is necessary to capture epistemic prediction uncertainty for a data acquisition metric. Using a Bayesian NN \cite{bishop_bayesian_1997, blundell_weight_2015} is one option for obtaining epistemic learning uncertainty. In a Bayesian NN, the weights are assigned probability distributions rather than fixed values. Therefore, the prediction output is not a point estimate but a Gaussian distribution. Bayesian NNs are trained through backpropagation, but they require significantly higher computational costs than conventional NNs to optimize the probability density distribution of weights and biases. 

Ensemble methods \cite{ying_zhao_survey_2005, yin_active_2007} combine multiple models to improve accuracy and provide uncertainty information. Accuracy improves more when the errors of the models are less correlated. In the extreme case where $M$ models have equal and entirely uncorrelated errors, averaging the models decreases error by a factor of $1/M$ \cite{fauser_neural_2015}. Error correlation decreases when models are less similar, which can be achieved by training on different subsets of data or using models with different underlying assumptions \cite{melville_diverse_2004}. 

Uteva et al. \cite{uteva_active_2018} tested an active learning scheme of sampling where two different GPR models disagreed the most. Because the GPR models were trained on different sets of data, they had different $\theta$-hyperparameters and thus different behavior. This method outperformed sampling the location of maximum predicted variance of a single GPR model. Lin et al. \cite{lin_automatically_2020} combined two NNs with different architectures into an ensemble, and added adaptive samples at the locations of maximum disagreement. Christiano et al. \cite{christiano_deep_2017} combined three neural networks in an ensemble and added additional training samples where the variance of the predictions was maximized. 

While existing ensemble methods can output a location of maximum variance for adaptive learning, they do not output a predictive probability distribution like GPR does. Such a predictive distribution is highly useful, offering additional insight into the model and enabling the use of probabilistic acquisition functions such as Expected Improvement. In this work, we present a formal statistical treatment to extract such a probability distribution. Specifically, we estimate the epistemic learning uncertainty by combining multiple E2NN models and calculating a Bayesian predictive distribution. 

Training a single E2NN model typically requires several minutes to reach convergence, which means updating the entire ensemble can be time-consuming. Therefore, a method combining Neural Networks and linear regression is explored to alleviate the cost of neural network training. Early exploration of the method was performed by Schmidt et al. \cite{schmidt_feedforward_1992}. The method combines two basic steps: first, creating a neural network with random weights and biases, and second, setting the last layer of weights using a linear regression technique such as ordinary least squares or ridge regression. These steps are far cheaper than backpropagation-based neural network training, enabling accelerated retraining of the E2NN ensemble during active learning. 

The next section, Section 2, contains a brief review of E2NN. Section 3 covers the proposed approach of adaptive and rapid learning, which is followed by Section 4 showing fundamental mathematical demonstrations and a practical aerospace engineering example involving predicting the aerodynamic performance of the generic hypersonic vehicle.

\section{Review of Emulator Embedded Neural Networks}

The E2NN has low fidelity models, called emulators, embedded in specific neurons of the neural network's architecture. The emulators contribute to regularization and preconditioning for increased accuracy. The emulators can take the form of any low-cost information source, such as reduced/decomposed physics-based models, legacy equations, data from a related problem, models with coarsened mesh or missing physics, etc.

\begin{figure}[h]
\centering
\includegraphics[width = 0.45\textwidth]{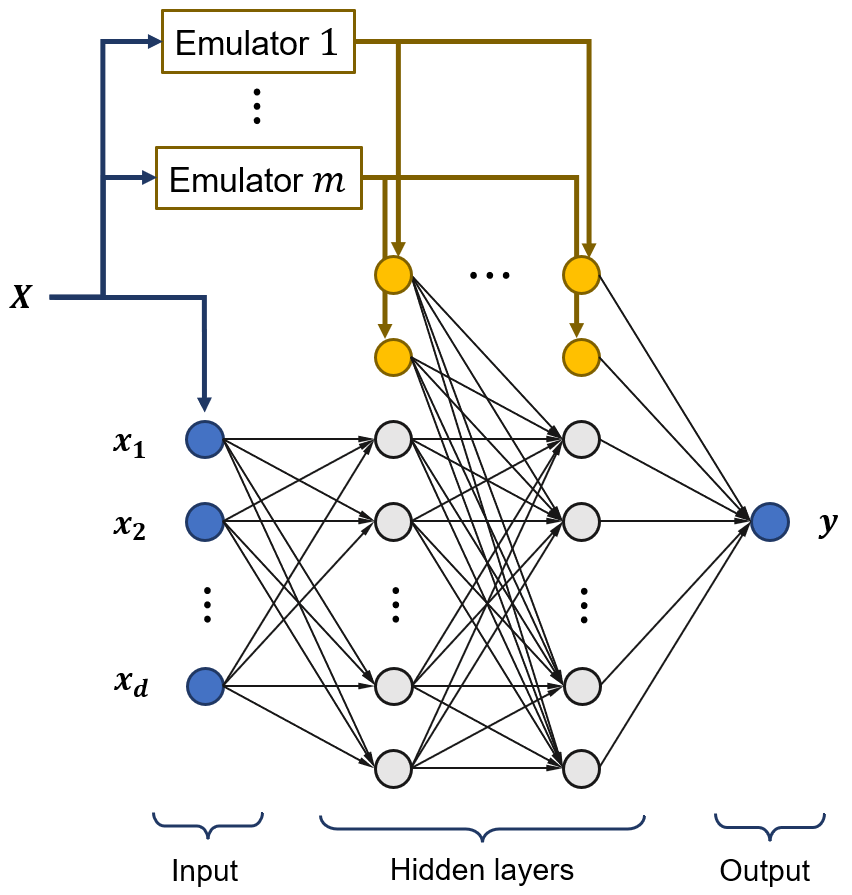}
\caption{Architecture of an Emulator Embedded Neural Network.}
\label{fig:e2nn}
\end{figure}

An emulator embedded in the last hidden layer can only be scaled before being added to the response, while an emulator embedded in the second-to-last hidden layer can be transformed through any functional mapping (by the Universal Approximation Theorem \cite{cybenko_approximation_1989, hornik_approximation_1991, leshno_multilayer_1993}) before being added to the response. The simplest solution when selecting the architecture is to embed the emulator in all hidden layers and allow the E2NN training to select which instances are useful. The flexibility of E2NN in selecting between an arbitrary number of LF models, each embedded in multiple hidden layers, enables wide applicability to problems with LF models of high, low, or unknown accuracy. The architecture of E2NN is illustrated in Fig. \ref{fig:e2nn}.

\section{Proposed Approach: Adaptive Learning with Non-Deterministic Emulator Embedded Neural Network}

The main technical contribution of this work is a novel method for combining predictions of an ensemble of models to approximate epistemic modeling uncertainty. This uncertainty information lowers training data costs by enabling Adaptive Learning (AL). For optimization problems, the Expected Improvement (EI) metric identifies adaptive sampling locations of maximum information gain. The EI assessment requires a probability density function, which is estimated by training an ensemble of E2NN models and estimating the underlying distribution of predictions. Greater disagreement among individual E2NN realizations is typically due to a lack of training samples and implies greater epistemic uncertainty. 

A second contribution involved speeding up E2NN training hundreds of times using Rapid Neural Network (RaNN) training, allowing fast model updating during AL iterations. Ensemble uncertainty estimation is discussed in Subsection \ref{subsection:uncertainty}, and AL using EI is explained in Subsection \ref{subsection:ei} Finally, the RaNN methodology is summarized in Subsection \ref{subsection:rann}, and practical considerations for avoiding numerical errors are discussed in Subsection \ref{subsection:numerical_errors}.

\subsection{Proposed Approach for Assessing Epistemic Modeling Uncertainty of E2NN}
\label{subsection:uncertainty}

The epistemic modeling uncertainty is estimated using an ensemble of E2NN models. The E2NN model predictions agree at the training points, but the predictions between points depend on the random initializations of the weights and biases. In other words, the E2NN model predictions are samples drawn from a stochastic prediction function in the design space. A higher magnitude of disagreement implies greater epistemic modeling uncertainty. This suggests a useful assumption: model the epistemic uncertainty as the aleatoric uncertainty of the E2NN model predictions. Specifically, assume the two pdfs are approximately equal at each prediction point $x$: 

\begin{equation}
    \label{eqn:pdfs}
    p(y_{true}(x)) \approx p(y_{E2NN}(x))
\end{equation}

However, finding the exact aleatoric distribution of $y_{E2NN}(x)$ requires training an infinite number of E2NN models. Instead, we use a finite number of model predictions at a point $x$ as data $D(x)$ to construct a posterior predictive distribution (ppd) for the E2NN predictions. This ppd is then used as an estimate of the epistemic pdf of the true function. 

\begin{equation}
    \label{eqn:pdf_est}
    p(y_{true}(x)) \approx p(y_{E2NN}(x)|D(x))
\end{equation}

Finding the ppd requires introducing a second assumption: The aleatoric E2NN predictions are normally distributed at each point $x$

\begin{equation}
    \label{eqn:e2nn_norm}
    y_{E2NN}(x) \sim N(\mu, \sigma)
\end{equation}

\begin{figure}[h]
\centering
\includegraphics[width = 0.6\textwidth]{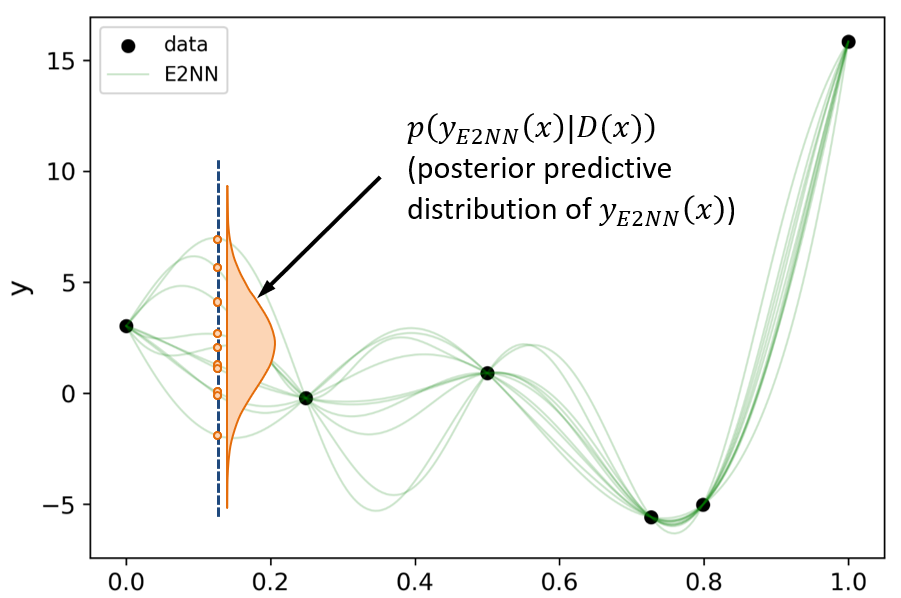}
\caption{Illustration of using multiple E2NN model realizations to estimate the underlying aleatoric probability distribution. This estimate is used as an approximation of the epistemic modeling uncertainty.}
\label{fig:ensemble}
\end{figure}

The process of combining multiple E2NN model predictions to estimate the epistemic uncertainty is illustrated in Fig. \ref{fig:ensemble}. 

For the general case of finding a posterior predictive distribution for a normal distribution from which we have $n$ iid data points $D$ but no additional information, we include a proof in Appendix \ref{section:derivation}. First, we need a prior and a likelihood. The prior for the mean and variance is a normal-inverse-chi-squared distribution ($NI\chi^2$) 

\begin{equation}
    p(\mu, \sigma^2) = NI\chi^2(\mu_0,\kappa_0,\nu_0,\sigma_0^2)=N(\mu|\mu_0, \sigma^2/\kappa_0) \cdot \chi^2(\sigma^2|\nu_0,\sigma_0^2)
\end{equation}

Here $\mu_0$ is the prior mean and $\kappa_0$ is the strength of the prior mean, while $\sigma_0^2$ is the prior variance and $\nu_0$ is the strength of the prior variance. The likelihood of the $n$ iid data points is the product of the likelihoods of the individual data points. By Bayes' rule, the joint posterior distribution for $\mu$ and $\sigma^2$ is proportional to the prior times the likelihood. Selecting the constants $\kappa_0$, $\sigma_0$ and $\nu_0$ for an uninformative prior and integrating to marginalize out $\mu$ and $\sigma$ in the posterior yields the predictive posterior distribution. Given $n$ iid samples, this is given by a t-distribution

\begin{equation}
    \label{eqn:posterior-predictive-body}
    p(y|D) = t_{n-1}\left(\bar{y}, \frac{1+n}{n}s^2\right)
\end{equation}

\noindent where $\bar{y}$ is the sample mean and $s^2$ is the sample variance. 

\begin{equation}
    s^2 = \frac{1}{n-1}\sum_i(y_i-\bar{y})^2
\end{equation}

The final pdf of $y$ given the data $D$ is a t-distribution instead of a normal distribution because it combines both Bayesian epistemic and aleatoric uncertainty. The epistemic component of the uncertainty can be reduced by increasing the number of samples. As the number of samples or ensemble models $n$ approaches $\infty$, the pdf of Eq. \ref{eqn:posterior-predictive-body} will approach a normal distribution with the correct mean and standard deviation. For $n=2$ samples ($\nu=1$) this yields a Cauchy, or Lorentzian, distribution, which has tails so heavy that the variance $\sigma^2$ and mean $\mu$ are undefined \cite{sivia_data_2012}. For $n=3$ samples ($\nu=2$) the mean $\mu$ is defined but the variance $\sigma^2$ is infinite. Therefore, more than $3$ ensemble models should always be used to estimate the mean when an uninformative prior is used. 

Substituting Eq. \ref{eqn:posterior-predictive-body} into Eq. \ref{eqn:pdfs} yields the final estimate of the epistemic modeling uncertainty distribution,

\begin{equation}
    \label{eqn:ensemble-dist}
    p(y_{true}(x))=t_{n-1}\left( \bar{y}_{E2NN}(x), \frac{1+n}{n}\cdot s_{E2NN}(x) \right)^2
\end{equation}

\noindent where $\bar{y}_{E2NN}(x)$ is the sample mean prediction of the $n$ E2NN models in the ensemble, 

\begin{equation}
    \bar{y}_{E2NN}(x) = \frac{1}{n}\sum_i{y_{E2NN_i}(x)}
\end{equation}

\noindent and $s_{E2NN}(x)$ is the sample variance of E2NN model predictions in the ensemble. 

\begin{equation}
    s_{E2NN}(x)^2 = \frac{1}{n-1}\sum_i(y_{E2NN_i}(x)-\bar{y}_{E2NN}(x))^2
\end{equation}

Use of an uninformative prior results in conservative and robust estimation of the epistemic modeling uncertainty. 

\subsection{Adaptive Learning (AL) Using Expected Improvement}
\label{subsection:ei}

Adaptive learning allows data collection to be focused in areas of interest and areas with high uncertainty, reducing the number of samples required for design exploration and alleviating computational costs. Typically, an acquisition function is used to measure the value of information gained from adding data at a new location. The overall process requires a few simple steps: 

\begin{enumerate}
    \item Generate HF responses from an initial design of experiments. 
    \item Use sample data to build an ensemble of E2NN models.
    \item Maximize the acquisition function using an optimization technique. 
    \item If the maximum acquisition value is above tolerance, add a training sample at the location and go to step 2. Otherwise, stop because the optimization has converged. 
\end{enumerate}

The criteria for where to add additional data depends on the design exploration goals. A common goal is global optimization. Global minimization seeks to find the minimum of a function in $d$-dimensional space.

\begin{equation}
    \label{eqn:minimization}
    x_{opt} = \argmin_{x \in \mathbb{R}^d} y(x)
\end{equation}

\begin{figure}[h]
\centering
\includegraphics[width = 0.65\textwidth]{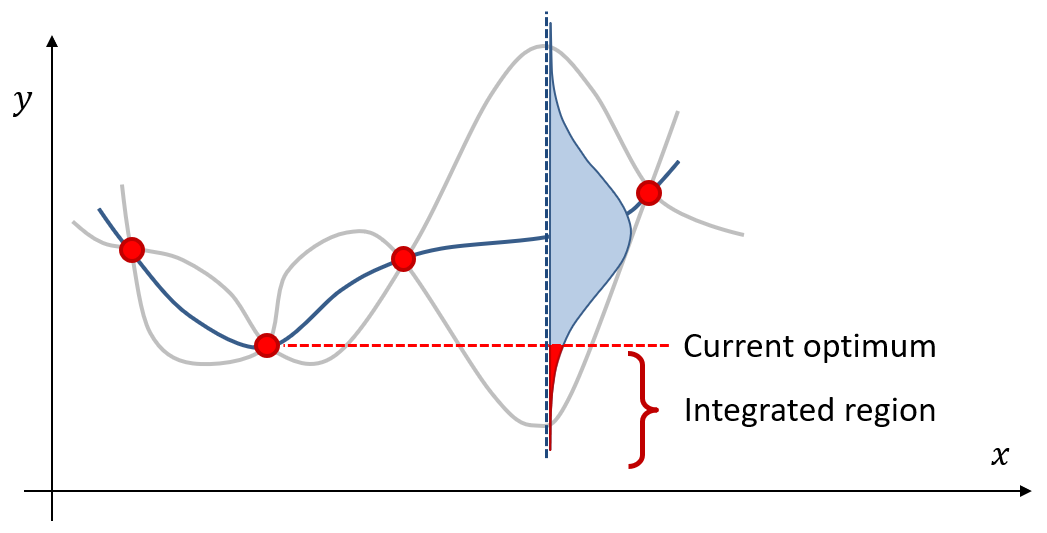}
\caption{Illustration of Expected Improvement calculation for adaptive learning.}
\label{fig:ei}
\end{figure}

For this objective, Expected Improvement (EI) is applicable \cite{jones_efficient_1998, beachy_expected_2020}. Informally, EI at a design point is the amount by which the design will, on average, improve over the current optimum. Formally, this is calculated by integrating the product of the degree and likelihood of every level of improvement. Likelihoods are given by the predictive probability distribution. Any new sample worse than the current optimum yields an improvement of 0. The general expression for EI is given by Eq. \ref{eqn:ei-integral} and illustrated in Fig. \ref{fig:ei}. 

\begin{equation}
    \label{eqn:ei-integral}
    EI(x) = \int_{-\infty}^{\infty} pdf(y)\cdot max(y_{current~opt}-y, 0)\cdot dy
\end{equation}

The EI value for a Gaussian predicted probability distribution is expressed as the closed-form expression,

\begin{equation}
    EI(x) = \left(f_{min}-\hat{y}(x)\cdot\Phi\left(\frac{f_{min}-\hat{y}(x)}{\sigma_z(x)}\right)+\sigma_z(x)\cdot\phi\left(\frac{f_{min}-\hat{y}(x)}{\sigma_z(x)}\right)\right)
\end{equation}

\noindent where $\phi(\cdot)$ and $\Phi(\cdot)$ are the pdf and cdf of the standard normal distribution, respectively; $\hat{y}(x)$ and $\sigma_z(x)$ are the mean and standard deviation of the predictive probability distribution, respectively; and $f_{min}$ is the current optimum. The current optimum can be defined as either the best sample point found so far, or the best mean prediction of the current surrogate. We use the former definition. The two definitions approach each other as the adaptive learning converges on the optimum. 

Unlike a kriging model, an E2NN ensemble returns a Student's t-distribution instead of a Gaussian distribution. The resulting formulation of EI in this case is \cite{tracey_upgrading_2018}

\begin{equation}
    \label{eqn:ei-t-dist}
    E[I(x)]=(f_{min}-\mu) \Phi_t(z) + \frac{\nu}{\nu-1}\left(1+\frac{z^2}{\nu}\right)\sigma\phi_t(z)
\end{equation}

\noindent where the t-score of the best point found so far is 

\begin{equation}
    \label{eqn:t-score}
    z = \frac{f_{min}-\mu}{\sigma}
\end{equation}

\noindent and $\phi_t(\cdot)$ and $\Phi_t(\cdot)$ are the pdf and cdf of the standard Student's t-distribution, respectively. Also, $\mu$, $\sigma$, and $\nu$ are the mean, scale factor, and degrees of freedom of the predictive t-distribution. 

\subsection{Training of E2NNs As Rapid Neural Networks}
\label{subsection:rann}

Typically, NNs are trained using auto-differentiation and backpropagation. The main problem in NN training is saddle points, which look like local minima because all gradients are zero and the objective function is increasing along almost all direction vectors. However, in a tiny fraction of possible directions the objective function decreases. If a NN has millions of weights, determining which direction to follow to escape from the saddle point is difficult. Optimizers such as Adam use momentum and other techniques to reduce the chance of getting stuck at a saddle point. However, a single E2NN model still takes significant computational time to reach convergence, requiring minutes to hours for a moderately sized neural network. Depending on the number of samples and the dimensionality of the problem, training an ensemble of E2NNs for adaptive learning will introduce significant computational costs.

To increase speed, the E2NN models can be trained as Rapid Neural Networks (RaNNs). Essentially, this involves initializing a neural network with random weights and biases in all layers and then training only the last layer connections. The last layer's weights and bias are trained by formulating and solving a linear regression problem such as ridge regression, skipping the iterative training process entirely. This accelerates training multiple orders of magnitude. These models are sometimes referred to as extreme learning machines, a term coined by Dr. Guang-Bin Huang, although the idea existed previously. 

An early example of RaNN by Schmidt et al. in 1992 utilized a feed-forward network with one hidden layer \cite{schmidt_feedforward_1992}. The weights connecting the input to the first layer were set randomly, and the weights connecting the hidden layer to the output were computed by minimizing the sum of squared errors. The optimal weights are analytically computable using standard matrix operations, resulting in very fast training. Saunders et al. \cite{saunders_ridge_1998} used ridge regression instead of least squares regression when computing the weights connecting the hidden layer to the output. 

Huang et al. later developed equivalent methods \cite{guang-bin_huang_extreme_2004,huang_extreme_2006}, and demonstrated that like conventional NNs, RaNNs are universal approximators \cite{huang_universal_2006}, with the ability to capture any continuous function over a bounded region to arbitrary accuracy if a sufficient number of neurons exist in the hidden layer. 

However, despite being universal approximators, RaNNs require many more hidden layer neurons to accurately approximate a complex function than NNs trained with backpropagation and gradient descent. This is because the hidden layer neurons are essentially functions that are linearly combined to yield the NN prediction. Backpropagation intelligently selects these functions, while RaNN relies on randomly chosen functions and therefore requires more of them to construct a good fit. If fewer neurons and higher robustness are desired for the final model, an ensemble of E2NN models can be trained using backpropagation after the adaptive learning has converged. 

In this work, we propose applying the random initialization and linear regression techniques to E2NN models instead of standard NNs. Multiple realizations of E2NN with RaNN training are combined into an ensemble, enabling uncertainty estimation and adaptive learning. The proposed framework of adaptive learning is demonstrated in multiple example problems in Section \ref{section:examples}.

\subsection{Practical Considerations for Avoiding Large Numerical Errors}
\label{subsection:numerical_errors}

Setting the last layer weights using linear regression sometimes causes numerical issues when capturing highly nonlinear functions, even when enough neurons are included. If ridge regression or LASSO is used, the E2NN model will not interpolate the training data points. If no regularization is used, the weights become very large to force interpolation. Large positive values are added to large negative values, resulting in round-off error of the E2NN prediction. The resulting fit is not smooth, but jitters up and down randomly. 

Numerical stability can be improved by using a Fourier activation function such as $\sin(x)$. This is reminiscent of a Fourier series, which can capture even non-continuous function to arbitrary accuracy. In fact, a NN with Fourier activation functions and a single hidden layer with $n$ neurons can capture the first $n$ terms of a Fourier series. Each neuron computes a different term of the Fourier Series, with the first layer of weights controlling frequency, the bias terms controlling offset, and the second layer weights controlling amplitudes. However, when using rapid training only the last layer weights are optimized. 

As points are added to a highly nonlinear function, interpolation becomes more difficult and numerical instability is introduced despite the Fourier activation. This is counteracted by increasing the frequency of the Fourier activation function, which enables more rapid changes of the fit. For smooth or benign functions, the Swish activation function tends to outperform Fourier. Therefore, we include some E2NN models using Swish and some using Fourier within the ensemble. Any model with any weights of magnitude above the tolerance of $100$ are considered unstable and dropped from the ensemble. Additionally, any model with NRMSE on the training data above the tolerance of $0.001$ is dropped from the ensemble, where NRMSE is defined as 

\begin{equation}
    NRMSE = \sqrt{\frac{\sum^N_{i=1}(\hat{y}_i-Y_i)^2}{\sum^N_{i=1}(\bar{Y}-Y_i)^2}}
\end{equation}

for predictions $\hat{y}_i$ and $N$ training samples $Y_i$. If over half of the Fourier models are dropped from the ensemble, all Fourier models have their frequencies increased and are retrained. These changes eliminate noisy and unstable models from the ensemble, as well as modifying models to remain stable as new points are added to the training data.


\section{Numerical Experiments}
\label{section:examples}

In aerospace applications, the costs of HF sample generation, i.e., computational fluid dynamics simulation, aeroelasticity analysis, coupled aerothermal analysis, etc., are typically far higher than the costs of generating LF samples and training NN models. Therefore, in the following examples, we compare the cost of various prediction models in terms of the number of HF samples required, rather than the computer wall-clock or GPU time. We assume that enough LF samples are collected to train an accurate metamodel, which is used to cheaply compute the emulator activations whenever the neural network makes predictions. 

In the following examples, we use a fully connected feed-forward neural network architecture. All LF functions are embedded as emulators in all hidden layers. The input variables are scaled to $[-1,1]$. The weights are initialized using the Glorot normal distribution. Biases are initialized differently for Swish and Fourier activation functions. Fourier biases are initialized uniformly between $[0,2\pi]$ to set the functions to a random phase or offset, while Swish biases are initialized uniformly on the region $[-4,4]$.

In total, each ensemble contains $16$ E2NN models with a variety of architectures and activation functions. Identical models make the same assumptions about underlying function behavior when deciding how to interpolate between points. If these assumptions are incorrect, the error will affect the whole ensemble. Therefore, including dissimilar models results in more robust predictions. 

Four activation functions and two different architectures yield $8$ unique NN models. Each unique model is included twice for a total of $16$ E2NN models in the ensemble. The four activation functions are $\swish(x)$, $\sin(\mathit{scale}~x)$, $\sin(1.1~\mathit{scale}~x)$, and $\sin(1.2~\mathit{ scale}~x)$. The two architectures include a small network and a large network. The small network has a single hidden layer with $2n$ neurons, where $n$ is the number of training samples. This means the number of neurons is dynamically increased as new training samples are added. The large network has two hidden layers, where the first hidden layer has $200$ neurons, and the second hidden layer has $5000$ neurons. Having most neurons in the second hidden layer enables more of the NN weights to be adjusted by linear regression. The large and small NNs use different scale terms for the Fourier activation functions. The large NN scale term is increased whenever more than half the large Fourier NNs have bad fits, and the small NN scale term is increased whenever over half the small Fourier NNs have bad fits. Because numerical instability is already corrected for, we do not use ridge regression. Instead, we perform unregularized linear regression using the Moore-Penrose inverse with a numerically stabilized $\Sigma$ term.

\subsection{One-dimensional analytic example with a linearly deviated LF model}

An optimization problem with the following form is considered.

\begin{equation}
    x_{opt} = \argmin_{x\in[0,1]} y_{HF}(x)
\end{equation}

Here $x$ is a design variable on the interval $[0,1]$. The high-fidelity function $y_{HF}(x)$ and its low-fidelity counterpart $y_{LF}(x)$ are given in Eqs. \ref{eqn:hf_1d} and \ref{eqn:lf_1d}. These functions have been used previously in the literature when discussing MF modeling methods \cite{bae_multifidelity_2020, forrester_recent_2009}. 

\begin{equation}
    \label{eqn:hf_1d}
    y_{HF}(x)=(6x-2)^2\sin(12x-4)
\end{equation}

\begin{equation}
    \label{eqn:lf_1d}
    y_{LF}(x)=0.5y_{HF}(x)+10(x-0.5)-5
\end{equation}

\begin{figure}[h]
\centering
\subfloat[Problem and initial samples]{\includegraphics[width=0.49\textwidth]{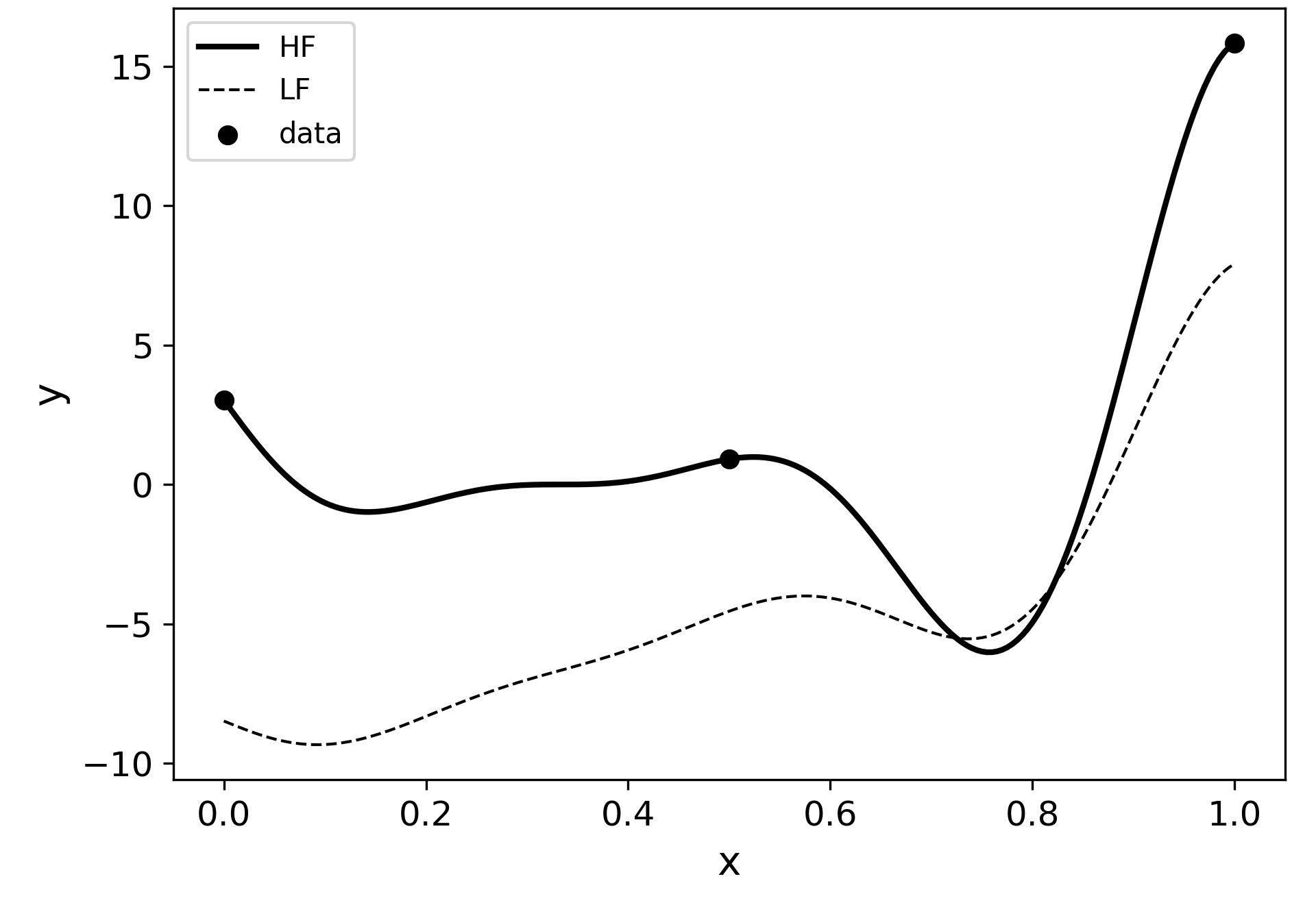}\label{fig:ex_1d_init_a}}
\hfill
\subfloat[Initial ensemble of E2NN models]{\includegraphics[width=0.49\textwidth]{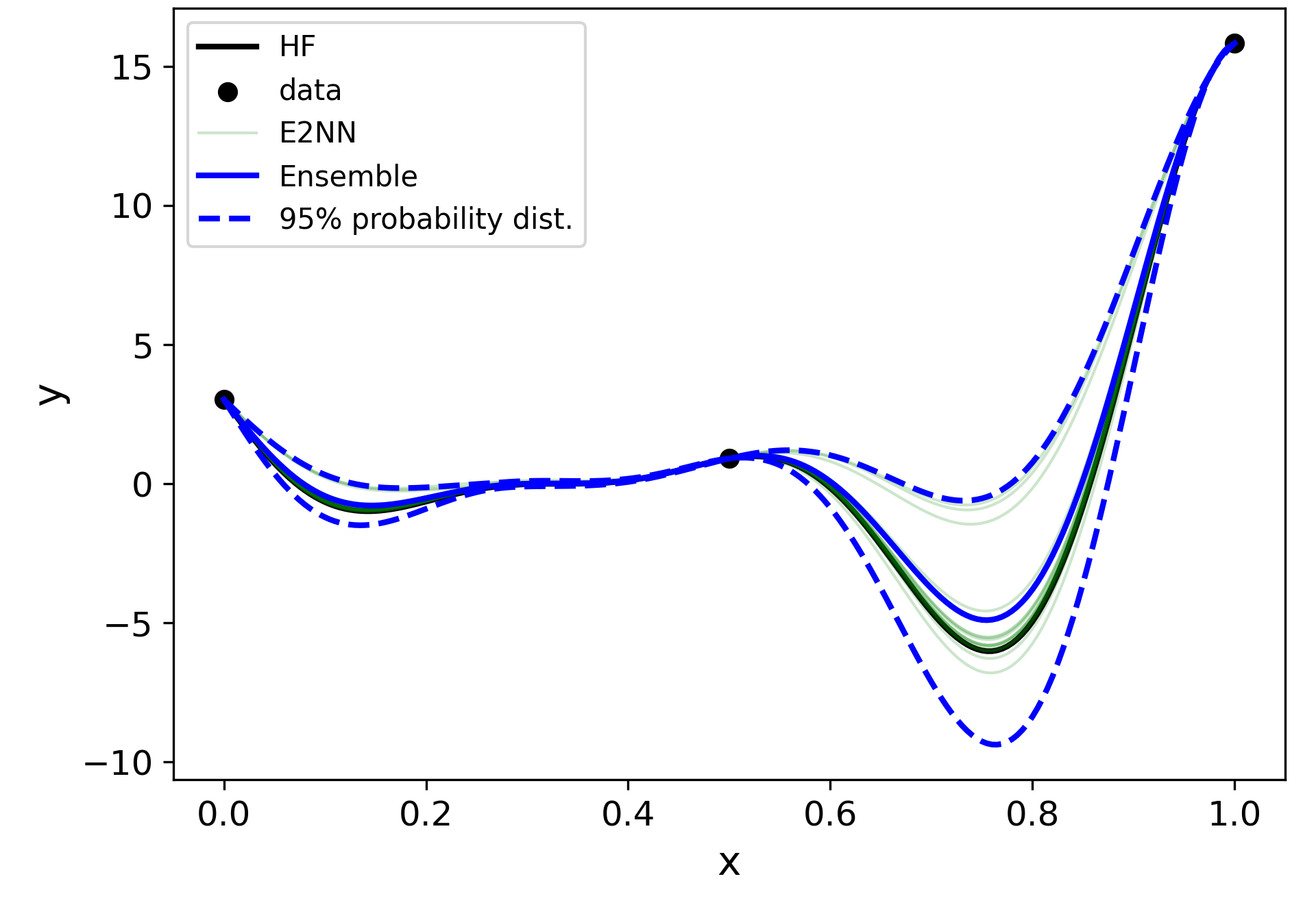}\label{fig:ex_1d_init_b}}
\caption{Initial problem and fitting of the E2NN model.}
\label{fig:ex_1d_init}
\end{figure}

As shown in Fig. \ref{fig:ex_1d_init_a}, the initial fit uses three HF samples at $x=[0,0.5,1]$ and an LF function which is linearly deviated from the HF function. The $16$ E2NN models used in the ensemble are shown in Fig. \ref{fig:ex_1d_init_b}. Three ensemble models are outliers, significantly overestimating the function value near the optimum. Two of these models nearly overlap, looking like a single model. All three of these inferior models are small NNs with Fourier activation functions. 

\begin{figure}[h]
\centering
\subfloat[Initial ensemble model]{\includegraphics[width=0.49\textwidth]{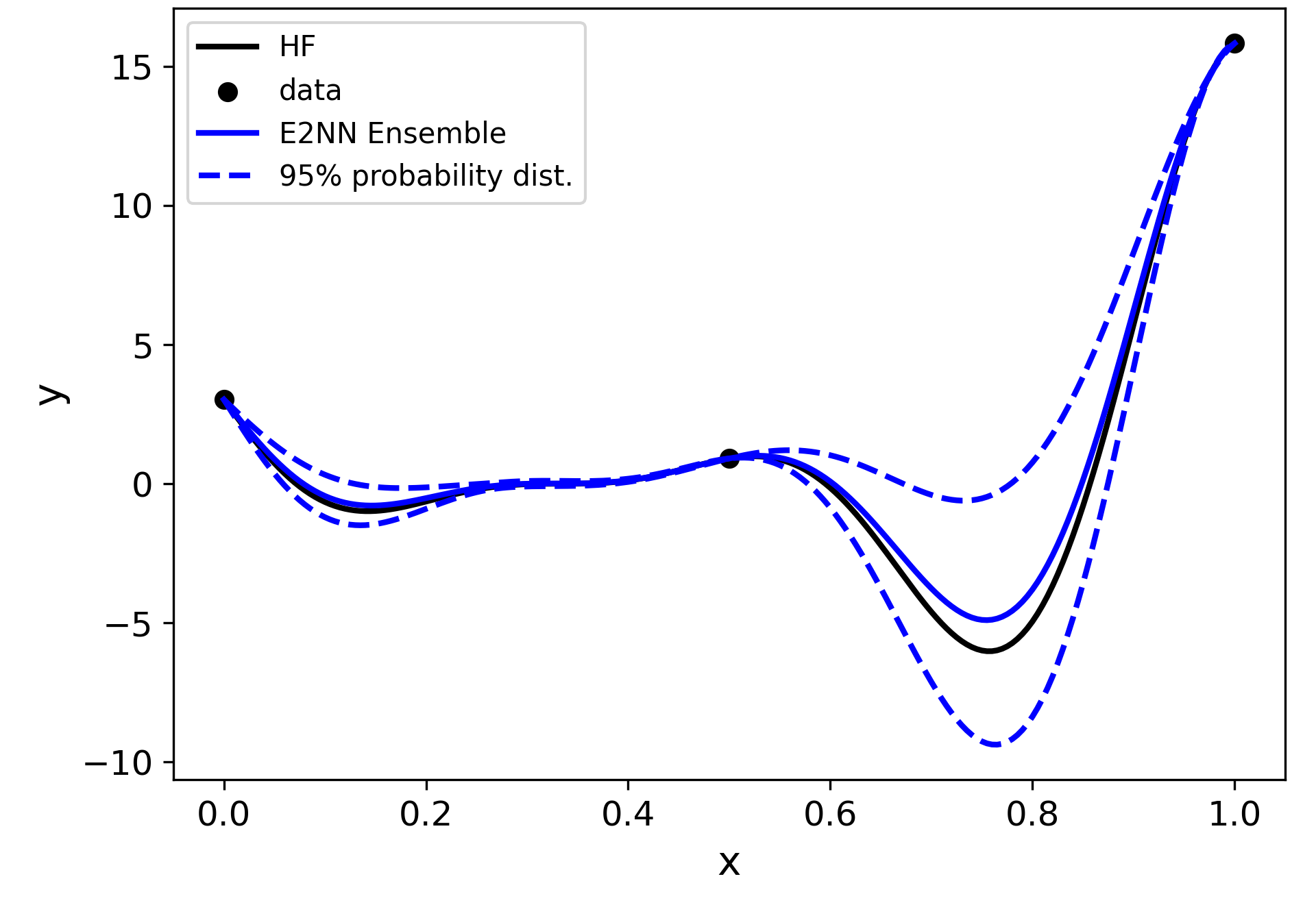}\label{fig:ex_1d_ei_a}}
\hfill
\subfloat[Expected Improvement of initial ensemble]{\includegraphics[width=0.49\textwidth]{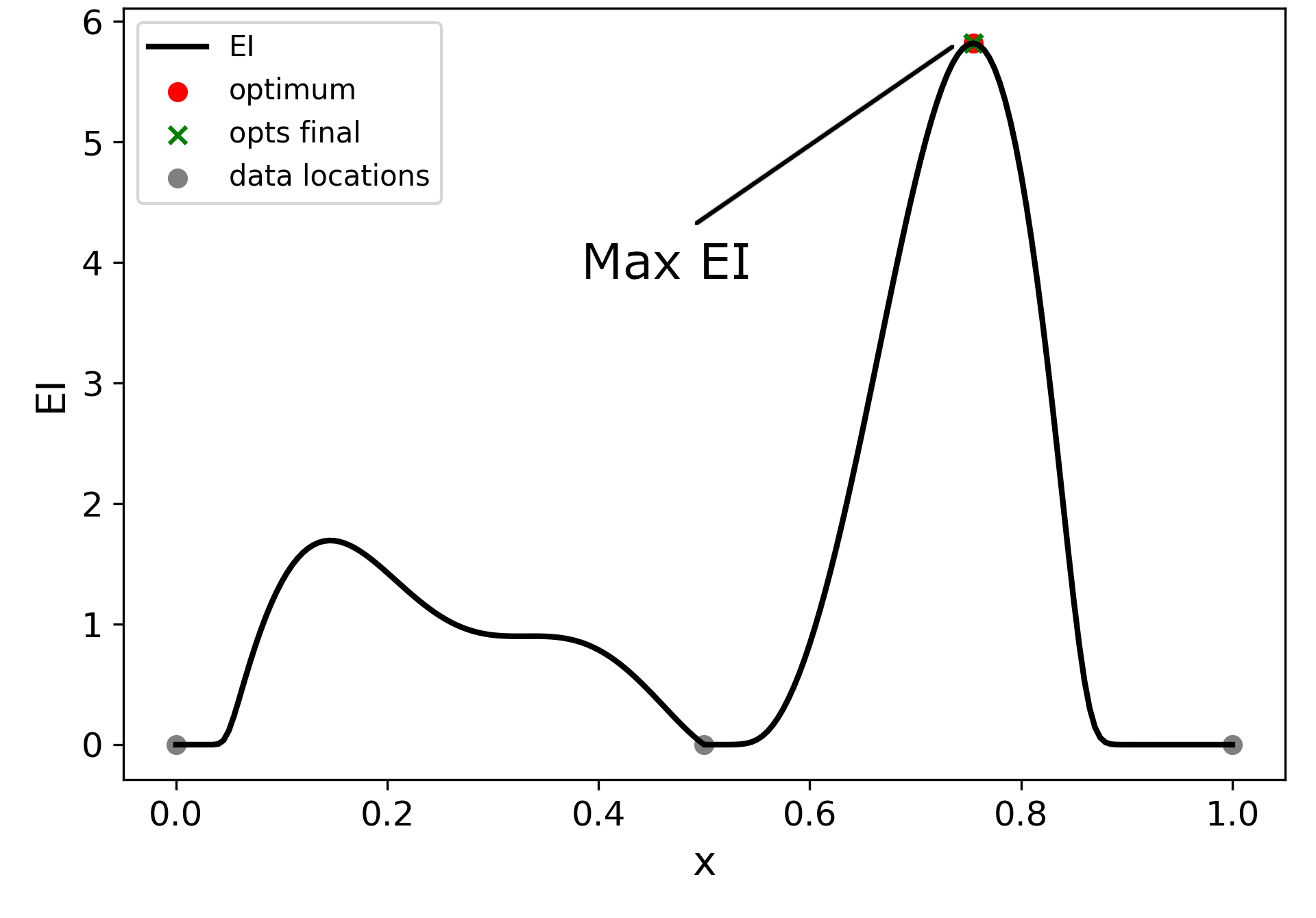}\label{fig:ex_1d_ei_b}}
\caption{Initial model and expected improvement.}
\label{fig:ex_1d_ei}
\end{figure}

The mean and $95\%$ probability range of the predictive t-distribution are both shown in Fig. \ref{fig:ex_1d_ei_a}. From this t-distribution, the Expected Improvement is calculated in Fig. \ref{fig:ex_1d_ei_b}. A new sample is added at the location of maximum EI. 

\begin{figure}[H]
\centering
\subfloat[First adaptive sample]{\includegraphics[width=0.49\textwidth]{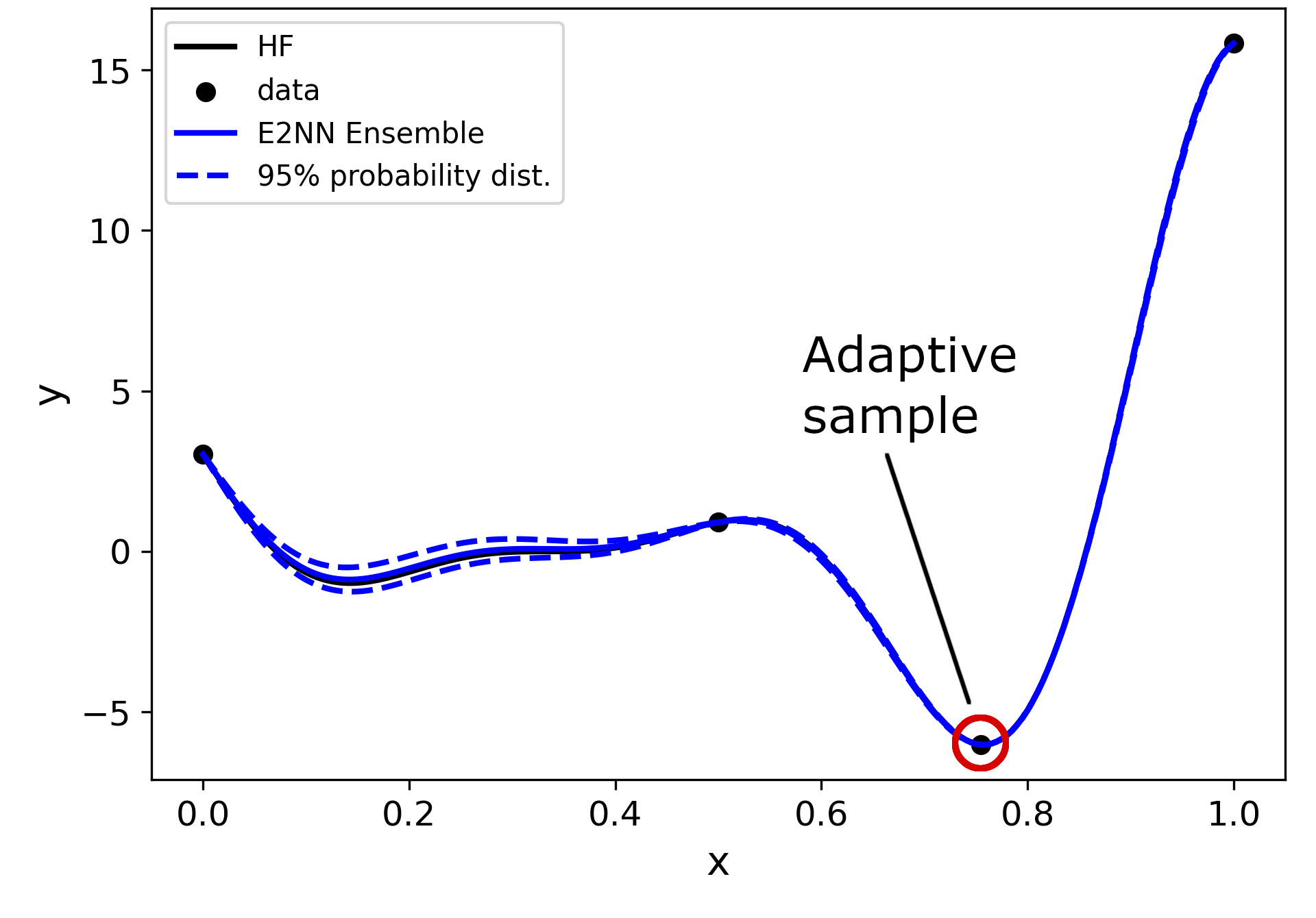}\label{fig:ex_1d_sample_a}}
\hfill
\subfloat[Second adaptive sample]{\includegraphics[width=0.49\textwidth]{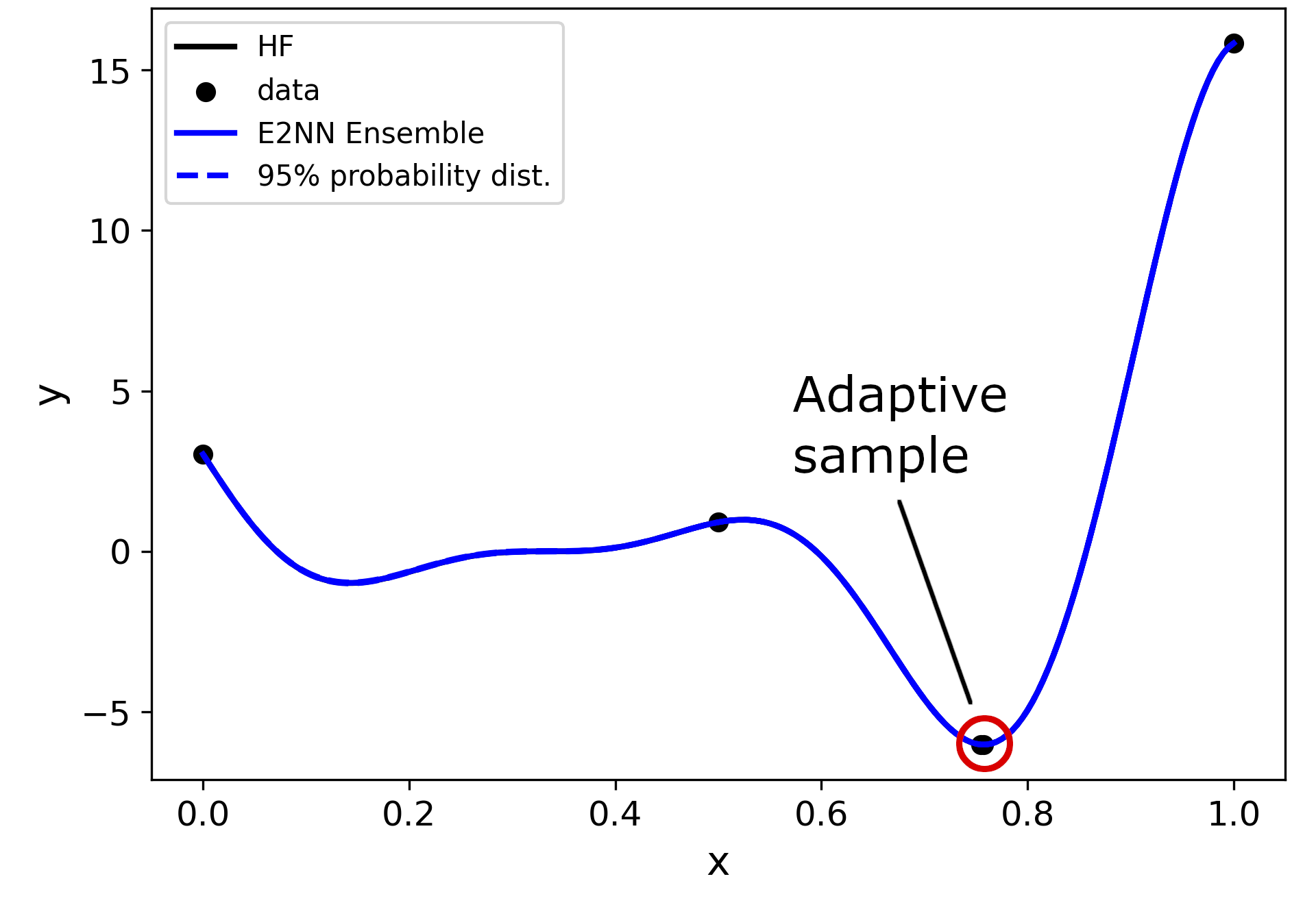}\label{fig:ex_1d_sample_b}}
\caption{Iterative model as adaptive samples are added.}
\label{fig:ex_1d_sample}
\end{figure}

The true optimum is $x_{opt}=0.7572$, $y_{opt}=-6.0207$. As shown in Fig. \ref{fig:ex_1d_sample_a}, the first adaptive sample at $x=0.7545$ lands very near the optimum, and the retrained ensemble's mean prediction is highly accurate. 

After the first adaptive sample, the maximum EI is still above tolerance. Therefore, an additional sample is added as shown in Fig. \ref{fig:ex_1d_sample_b}. The second adaptive sample at $x=0.7571$ is only $10^{-4}$ from the true optimum. After the second adaptive sample, the maximum EI value falls below tolerance, and the adaptive sampling converges.

\subsection{Two-dimensional analytical example}

The proposed ensemble method is compared with the popular kriging method for minimization of the following two-dimensional function.

\begin{equation}
    y_{HF}(x_1, x_2)=\sin(21(x_1-0.9)^4)\cos(2(x_1-0.9))+(x_1-0.7)/2+2*x_2^2\sin(x_1x_2)
\end{equation}

This nonstationary test function was introduced in \cite{rumpfkeil_construction_2017, clark_engineering_2016}[49,50]. The kriging method uses only HF training samples during optimization, but E2NN makes use of the following LF function. 

\begin{equation}
    y_{LF}(x_1,x_2)=\frac{y_{HF}(x_1,x_2)-2+x_1+x_2}{1+0.25x_1+0.5x_2}
\end{equation}

\begin{figure}[h]
\centering
\includegraphics[width = 0.5\textwidth]{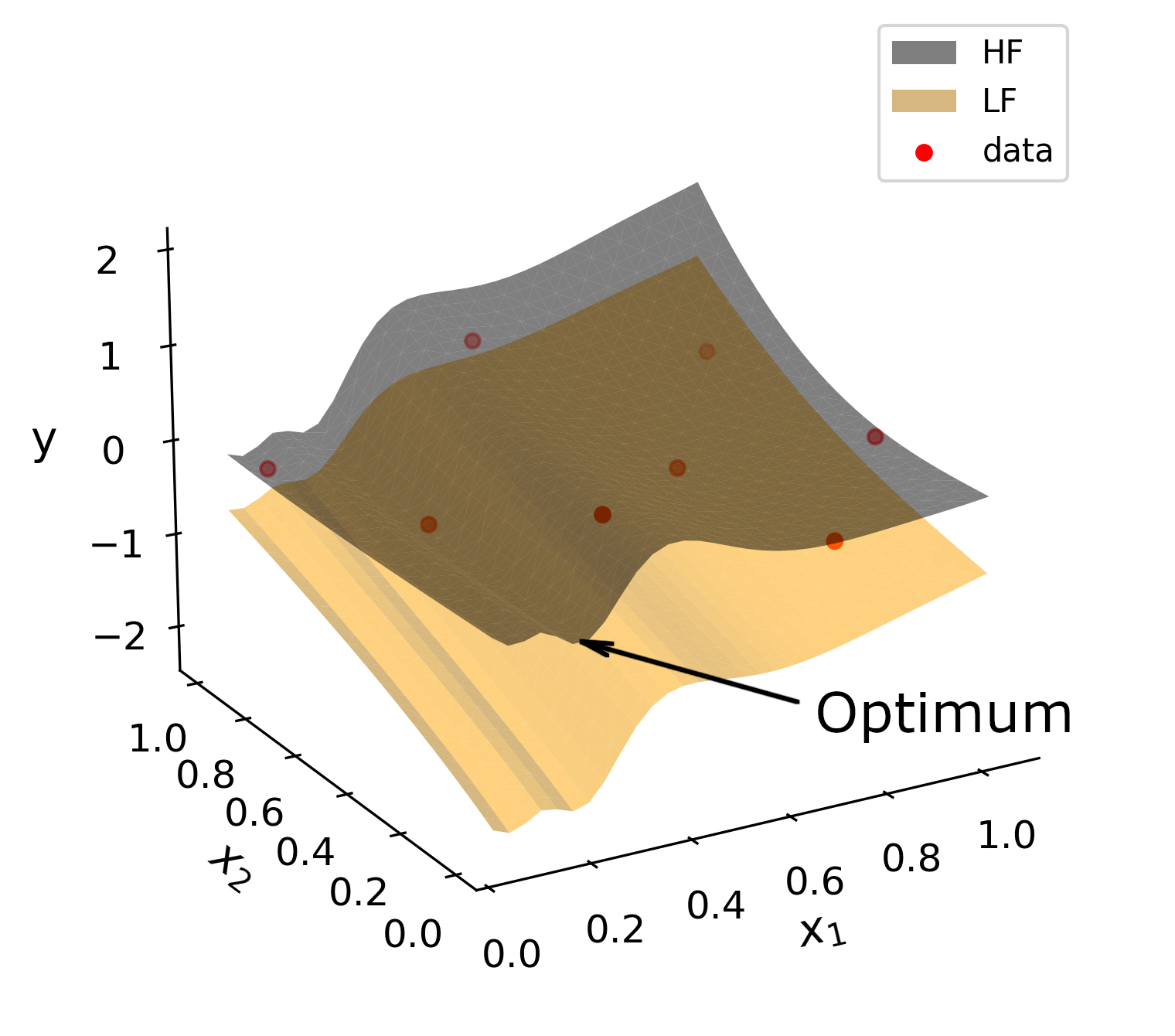}
\caption{Comparison of HF and LF functions for a nonstationary test function.}
\label{fig:ex_2d}
\end{figure}

The independent variables $x_1$ and $x_2$ are constrained to the intervals $x_1\in[0.05,1.05]$, $x_2\in[0,1]$. The LF function exhibits nonlinear deviation from the HF function as shown in Fig. \ref{fig:ex_2d}. 

\begin{figure}[h]
\centering
\subfloat[Initial Ensemble model (8 samples)]{\includegraphics[width=0.33\textwidth]{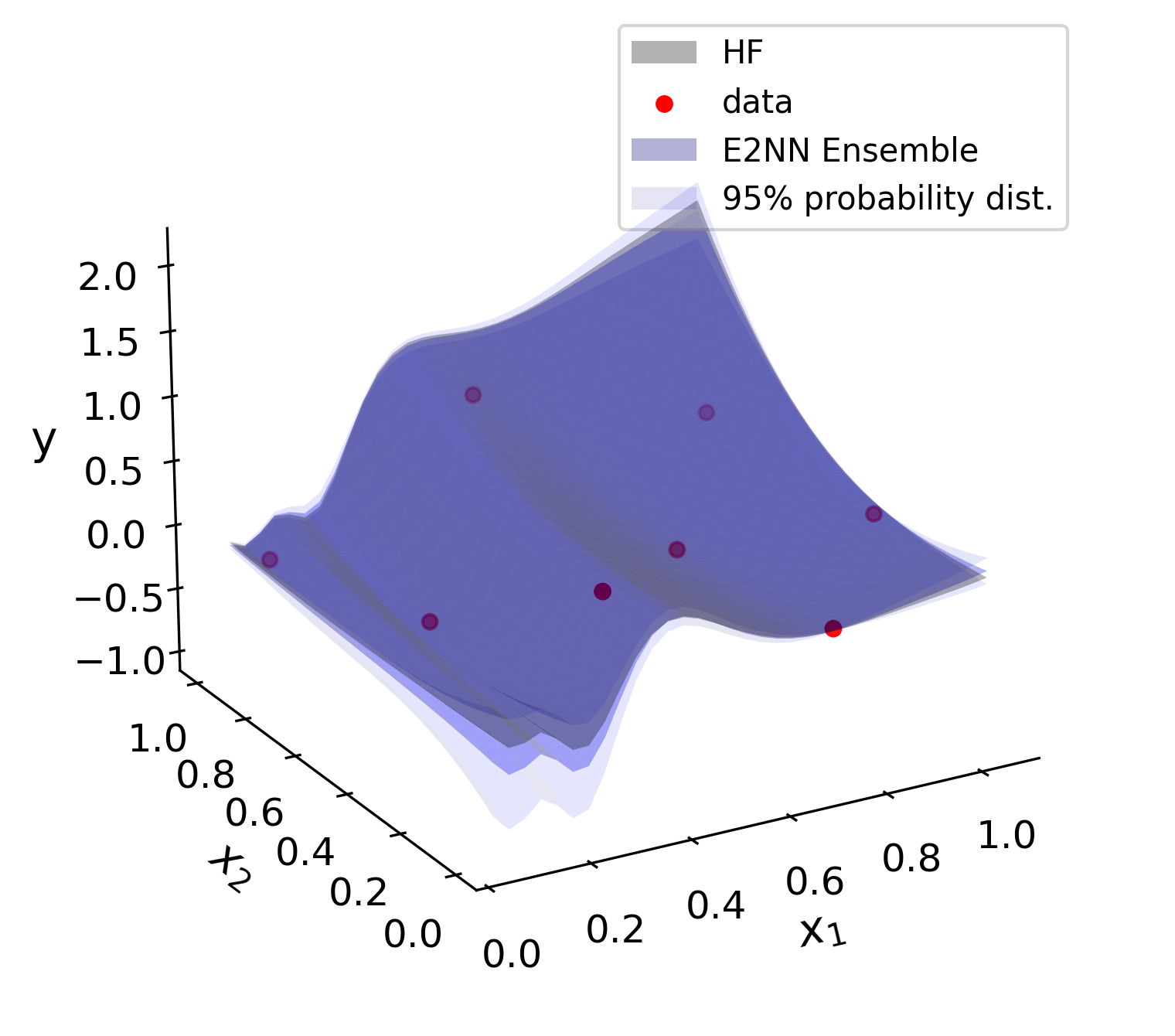}\label{fig:ex_2d_e2nn_a}}
\subfloat[Expected Improvement]{\includegraphics[width=0.33\textwidth]{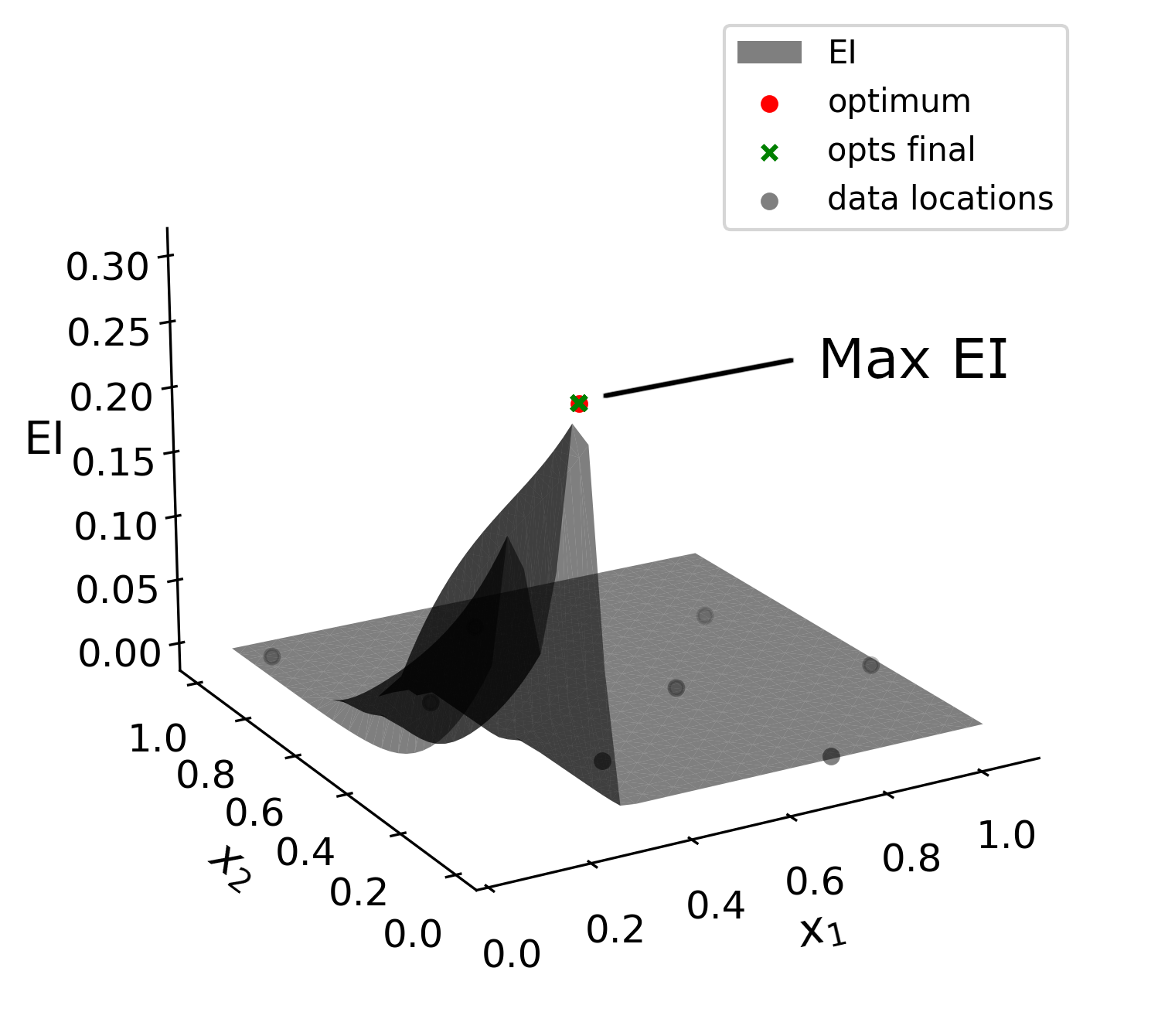}\label{fig:ex_2d_e2nn_b}}
\subfloat[Ensemble after 1 iteration (9 samples)]{\includegraphics[width=0.33\textwidth]{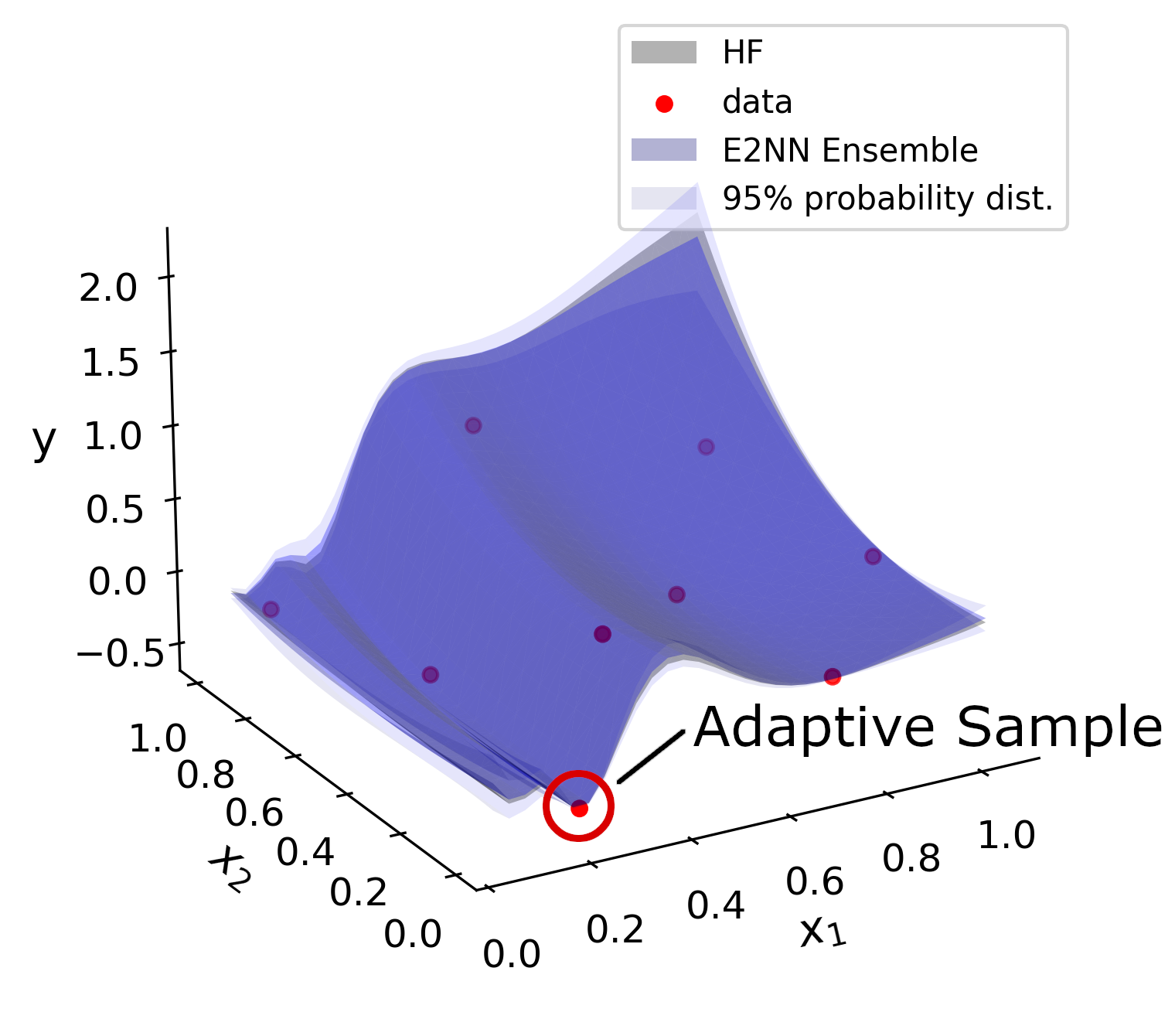}\label{fig:ex_2d_e2nn_c}}
\caption{Adaptive sampling of E2NN ensemble.}
\label{fig:ex_2d_e2nn}
\end{figure}

Eight training samples are selected to build the initial model using Latin hypercube sampling. The initial E2NN ensemble prediction is shown in Fig. \ref{fig:ex_2d_e2nn_a}. The resulting EI is shown in Fig. \ref{fig:ex_2d_e2nn_b}, and the ensemble prediction after a new sample is added is shown in Fig. \ref{fig:ex_2d_e2nn_c}. The initial fit is excellent, with only a small difference between the mean prediction and HF function. After the first adaptive sample is added, the optimum is accurately pinpointed. 

\begin{figure}[h]
\centering
\subfloat[Kriging with 28 samples]{\includegraphics[width=0.33\textwidth]{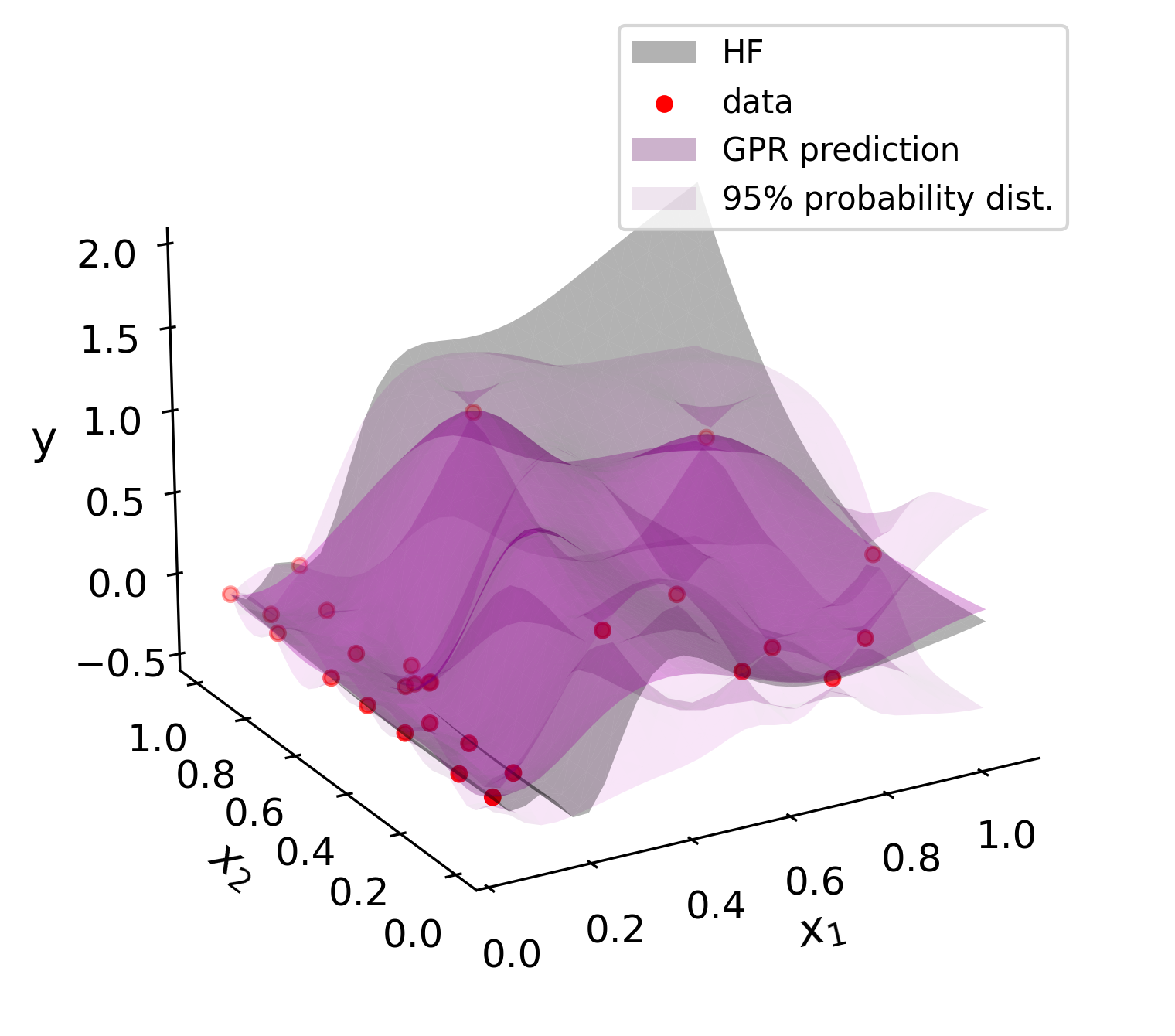}\label{fig:ex_2d_gpr_a}}
\subfloat[Expected Improvement]{\includegraphics[width=0.33\textwidth]{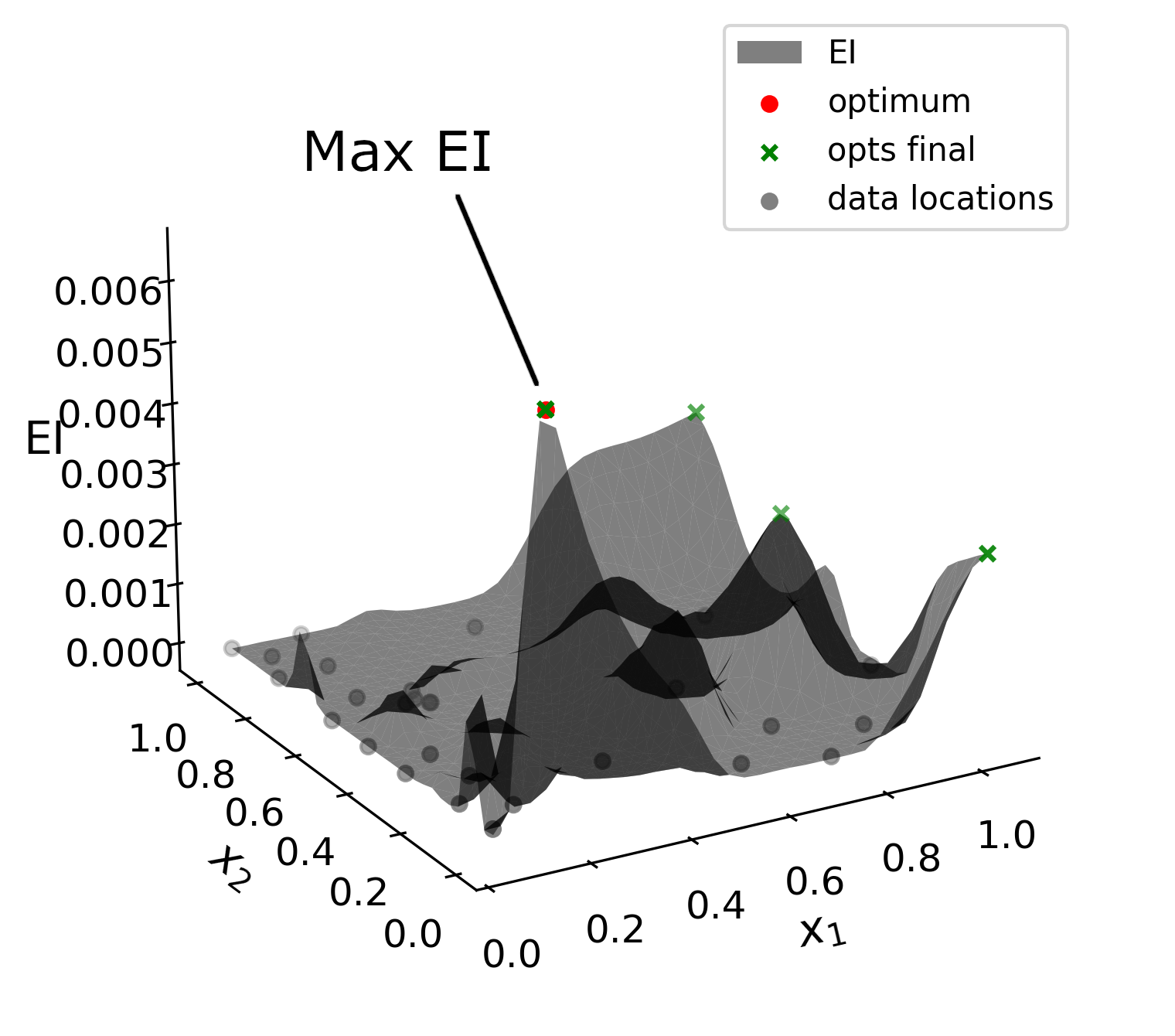}\label{fig:ex_2d_gpr_b}}
\subfloat[Kriging with 29 samples]{\includegraphics[width=0.33\textwidth]{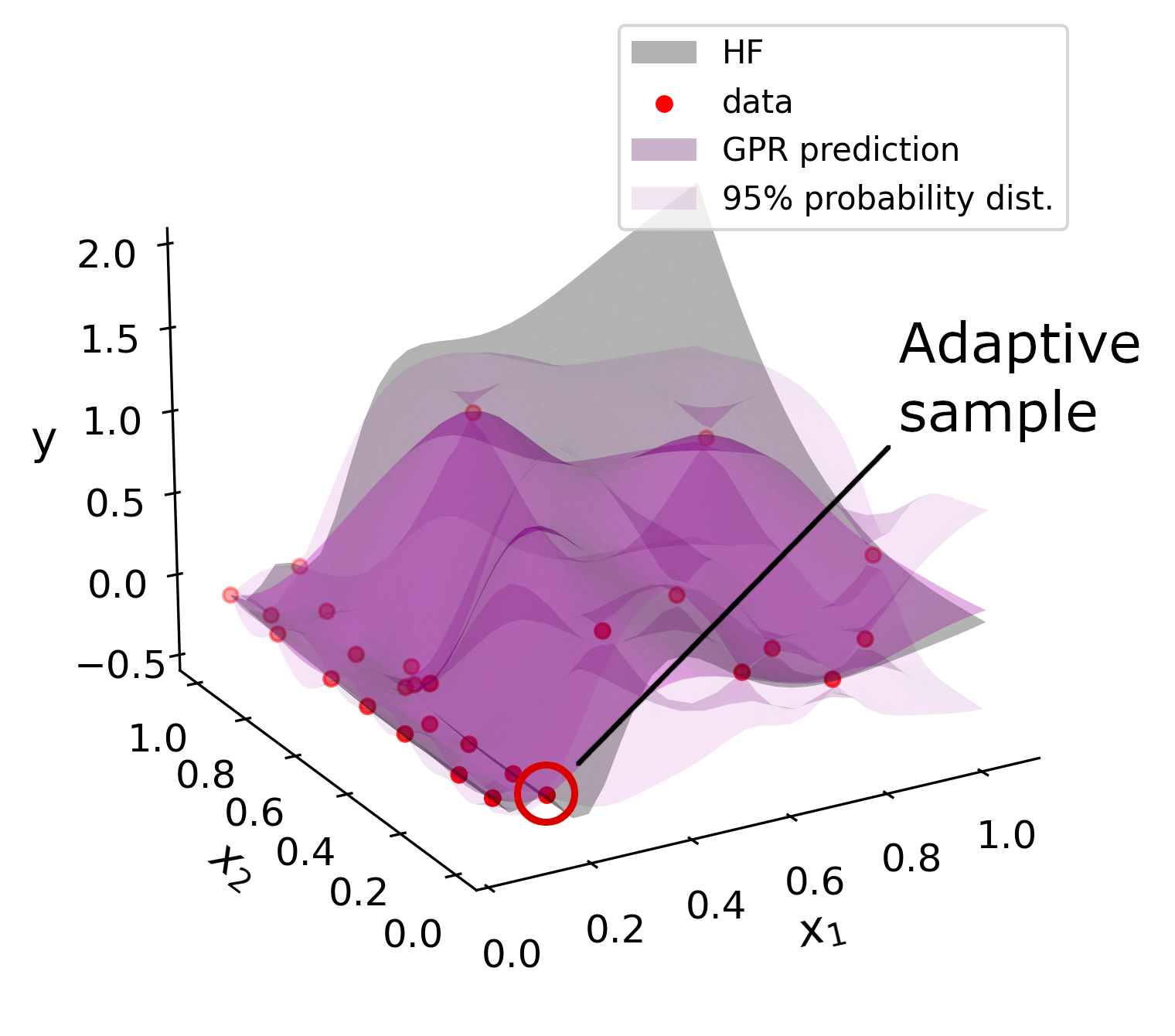}\label{fig:ex_2d_gpr_c}}
\caption{Adaptive sampling of kriging model.}
\label{fig:ex_2d_gpr}
\end{figure}

To compare the performance of adaptive learning with single fidelity kriging, EGO with a kriging model is run with the same initial set of $8$ points. The best kriging sample is shown in Fig. \ref{fig:ex_2d_gpr_a}. After adding a sample near the location of the optimum, the kriging model still does not capture the underlying trend of the HF model, as shown in Fig. \ref{fig:ex_2d_gpr_c}. The E2NN ensemble maintains higher accuracy over the design space by leveraging the LF emulator. 

\begin{figure}[H]
\centering
\subfloat[Converged ensemble model (11 samples)]{\includegraphics[width=0.33\textwidth]{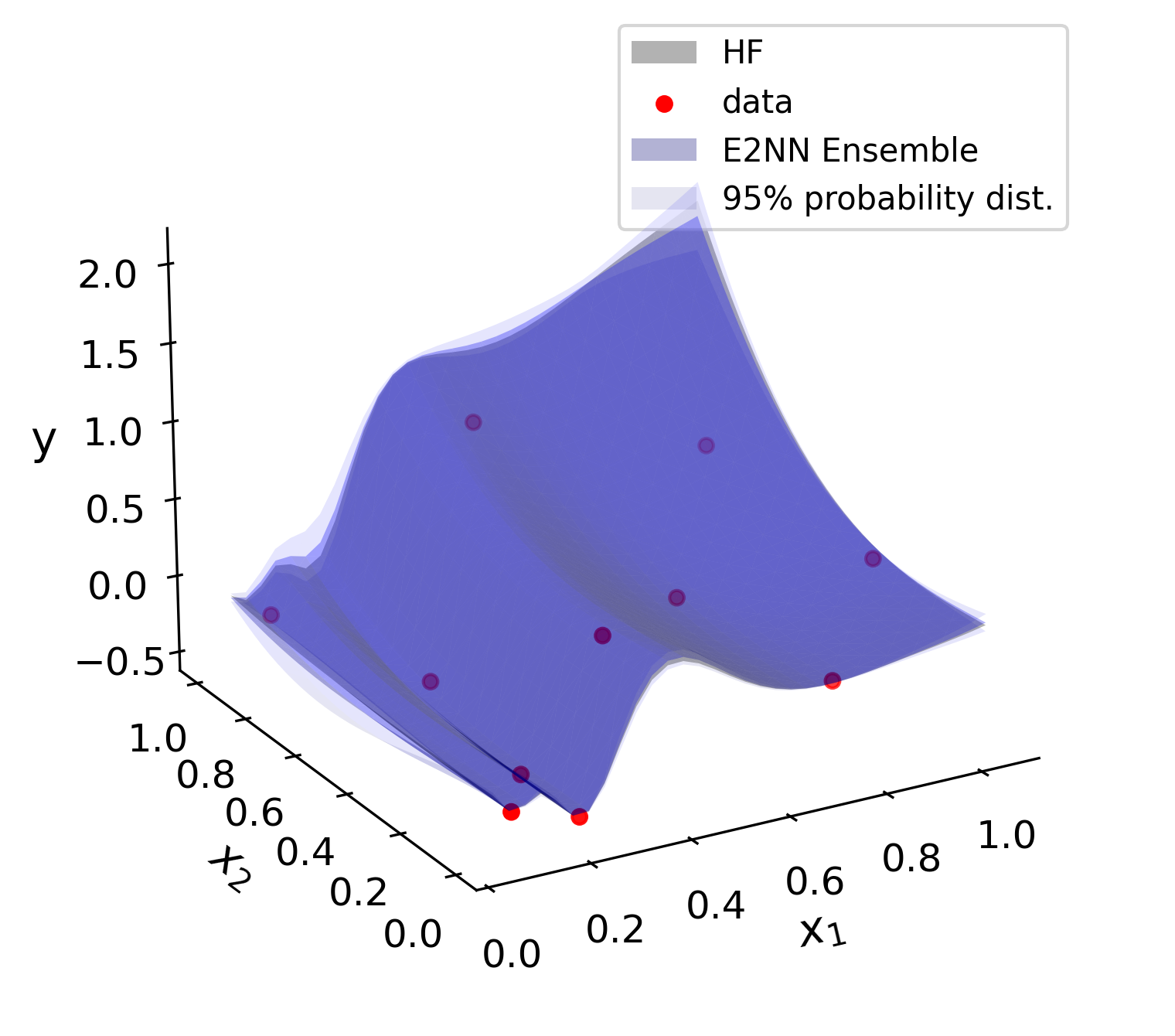}\label{fig:ex_2d_done_a}}
\hspace{0.1in}
\subfloat[Converged kriging model (37 samples)]{\includegraphics[width=0.33\textwidth]{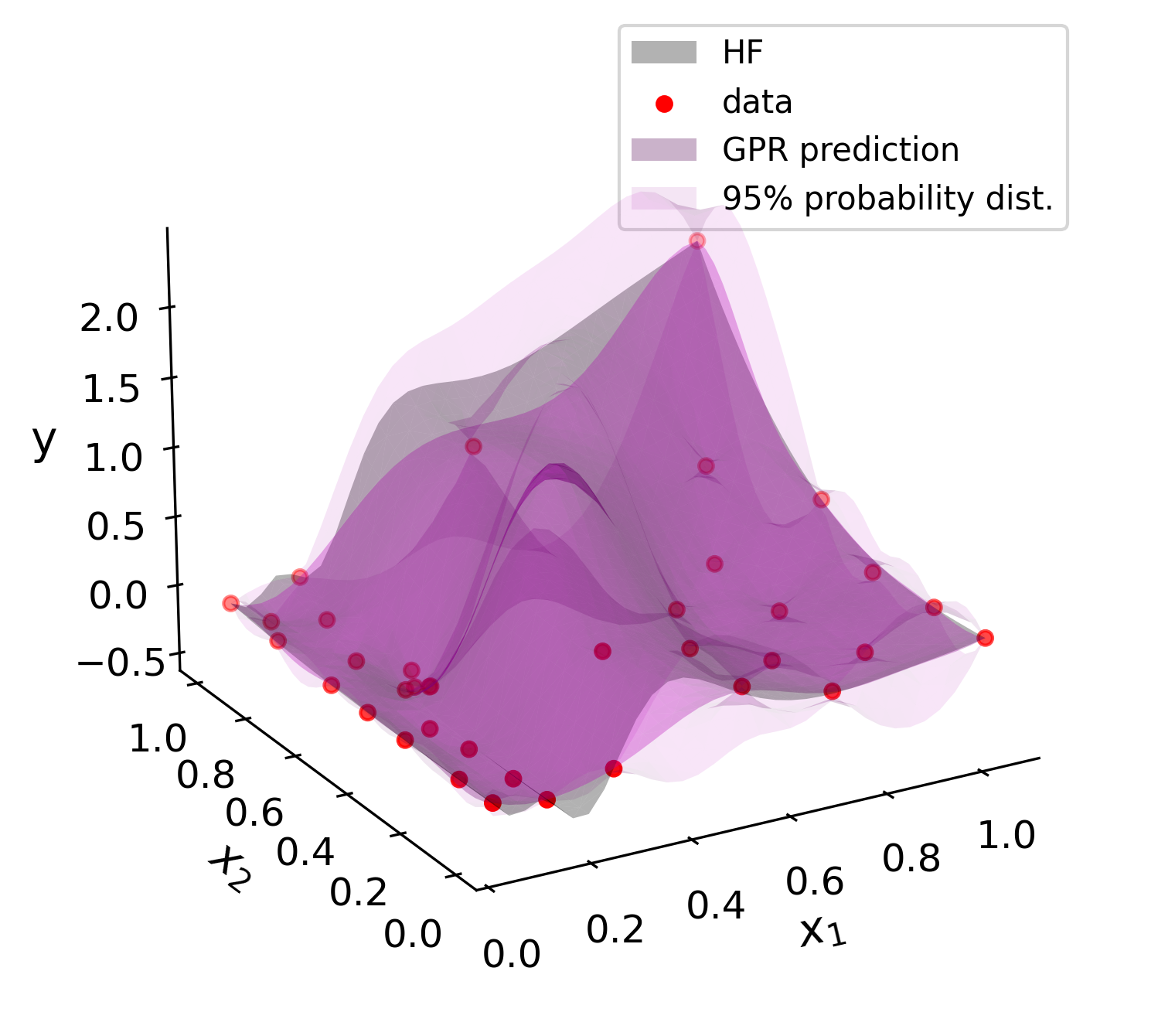}\label{fig:ex_2d_done_b}}
\caption{Comparison of final converged E2NN ensemble and kriging models.}
\label{fig:ex_2d_done}
\end{figure}

For the E2NN ensemble, the algorithm adds one more sample further up the valley that the optimum lies in, and then terminates because the expected improvement converges below tolerance. The final fit is shown in Fig. \ref{fig:ex_2d_done_a}. The kriging model adds $29$ samples before it converges, and still doesn't find the exact optimum, as shown in Fig. \ref{fig:ex_2d_done_b}. 

\subsection{Three-dimensional CFD example using a Hypersonic Vehicle Wing} 

\begin{figure}[h]
\centering
\subfloat{\includegraphics[width=0.50\textwidth]{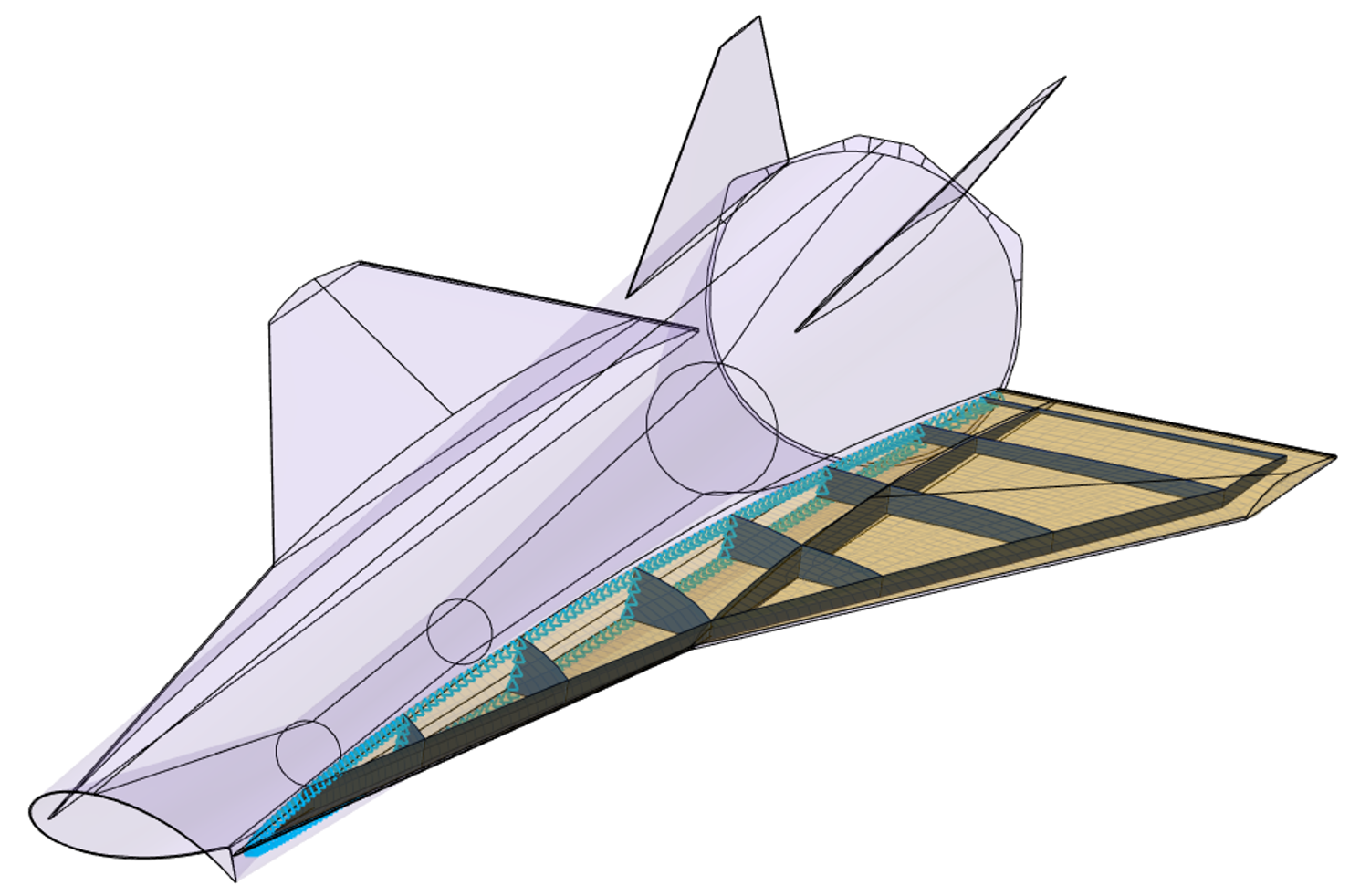}\label{fig:wing_a}}
\hfill
\subfloat{\includegraphics[width=0.44\textwidth]{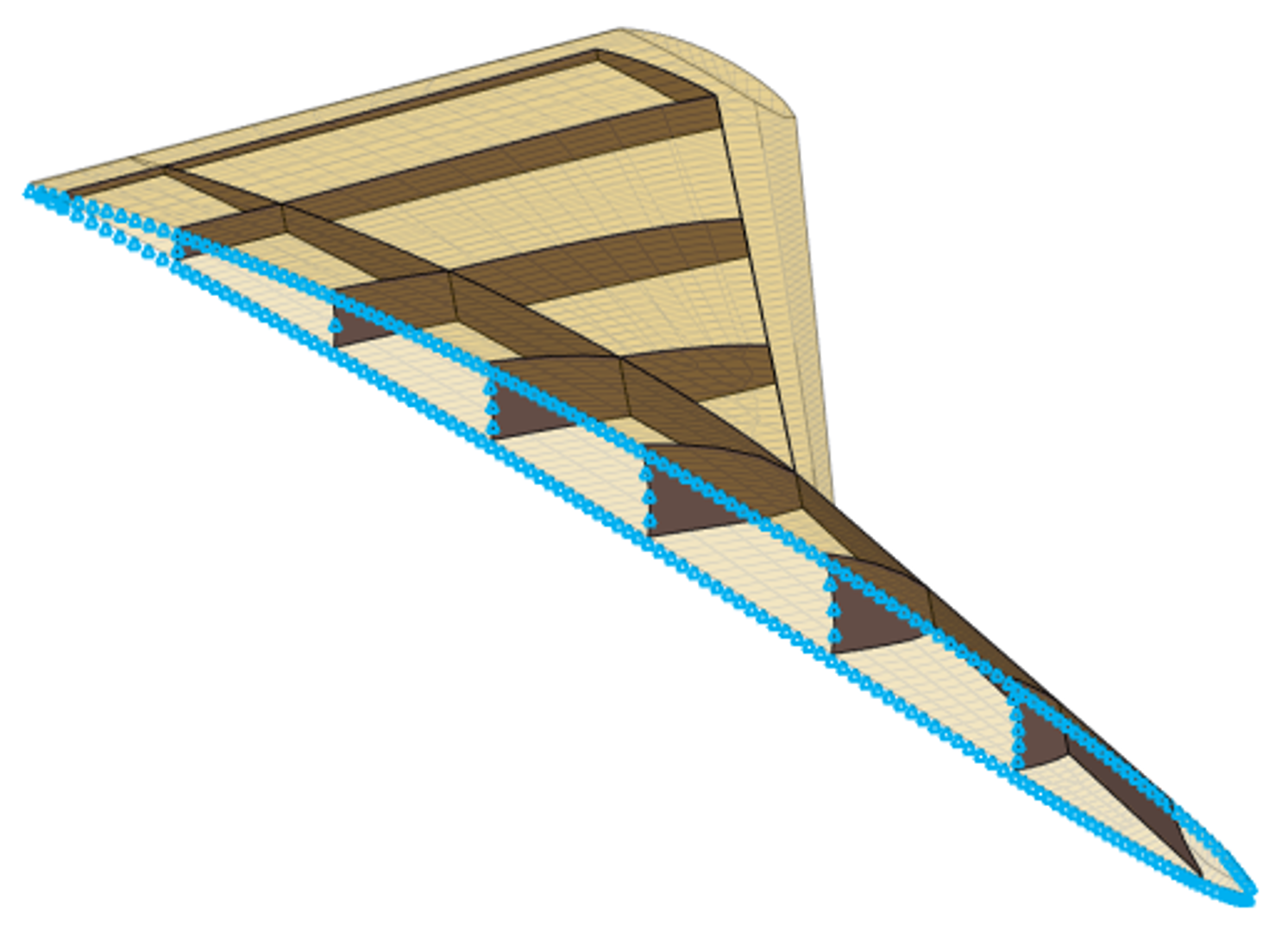}\label{fig:wing_b}}
\caption{Illustration of GHV wing used in CFD analysis.}
\label{fig:wing}
\end{figure}

This example explores modeling the Lift to Drag Ratio ($CL/CD$) of a wing of the Generic Hypersonic Vehicle (GHV) given various flight conditions. The GHV was developed at the Wright-Patterson Air Force Base to allow researchers outside the Air Force Research Laboratory to perform hypersonic modeling studies. The parametric geometry of the wing used was developed by researchers at the Air Force Institute of Technology \cite{boyd_generic_2020} and is shown in Fig. \ref{fig:wing}.

For this example, we studied the maximum lift-to-drag ratio of the GHV wing design with respect to three operational condition variables: Mach number (normalized as $x_1$), Angle of Attack (normalized as $x_2$) and Altitude (normalized as $x_3$). The Mach number ranges from $[1.2,4.0]$, while the angle of attack ranges from $[-5^\circ,8^\circ]$ and the altitude ranges from $[0, 50 \text{ km}]$. The speed of sound decreases with altitude, so the same Mach number denotes a lower speed at higher altitude. Atmospheric properties were calculated using the scikit-aero python library based on the U.S. 1976 Standard Atmosphere. 

\begin{figure}[h]
\centering
\subfloat[LF mesh with 30,000 cells]{\includegraphics[height=0.34\textwidth]{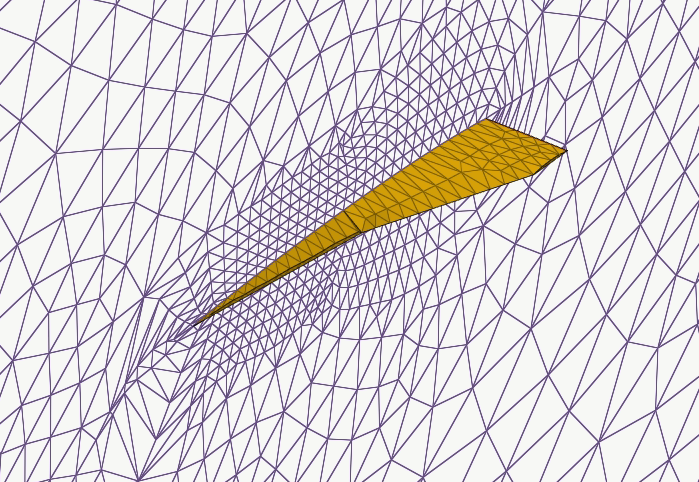}\label{fig:mesh_a}}
\hspace{0.1in}
\subfloat[HF mesh with 272,000 cells]{\includegraphics[height=0.34\textwidth]{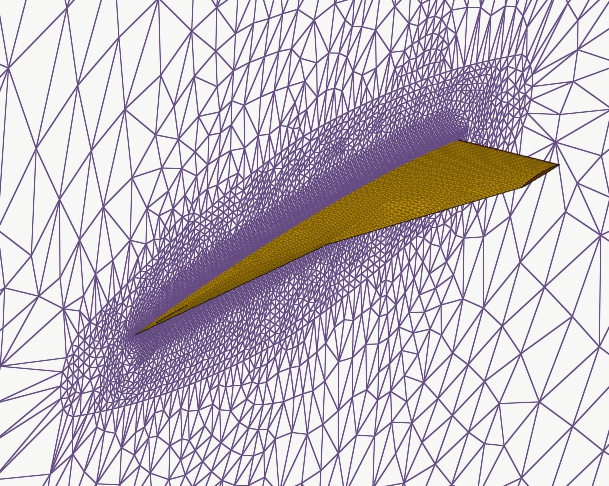}\label{fig:mesh_b}}
\caption{Comparison of HF and LF meshes used in CFD analysis.}
\label{fig:mesh}
\end{figure}

FUN3D \cite{anderson_implicit_1994} performed the CFD calculations. We used Reynolds Averaged Navier Stokes (RANS) with a mesh of $272,007$ tetrahedral cells for the HF model, and Euler with a mesh of $29,643$ tetrahedral cells for the LF model. To enable rapid calling of the LF model during each NN evaluation, $300$ evaluations of the LF model were performed and used to train a GPR model. This GPR model was then used as the LF function. The HF and LF meshes are compared in Fig. \ref{fig:mesh}.

\begin{figure}[h]
\centering
\subfloat[Mach ($x_1$) and AoA ($x_2$)]{\includegraphics[height=0.28\textwidth]{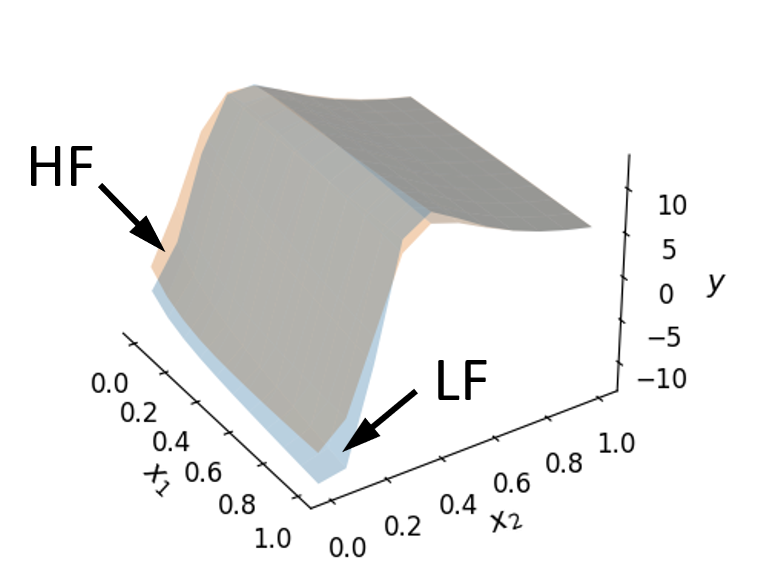}\label{fig:sweeps_a}}
\subfloat[Mach ($x_1$) and Altitude ($x_3$)]{\includegraphics[height=0.28\textwidth]{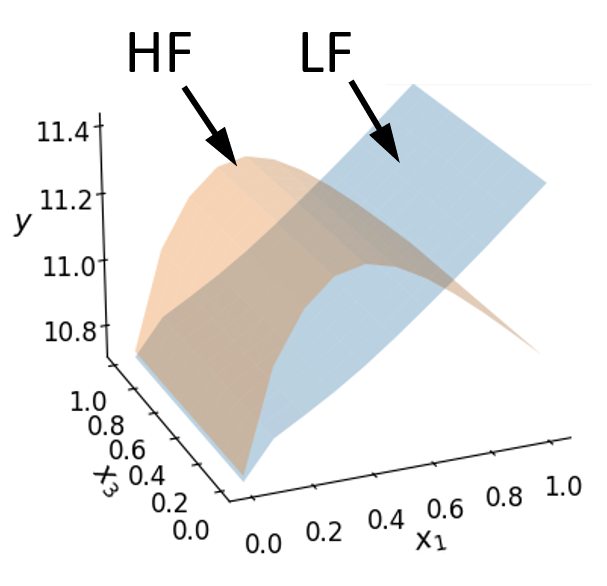}\label{fig:sweeps_b}}
\subfloat[AoA ($x_2$) and Altitude ($x_3$)]{\includegraphics[height=0.28\textwidth]{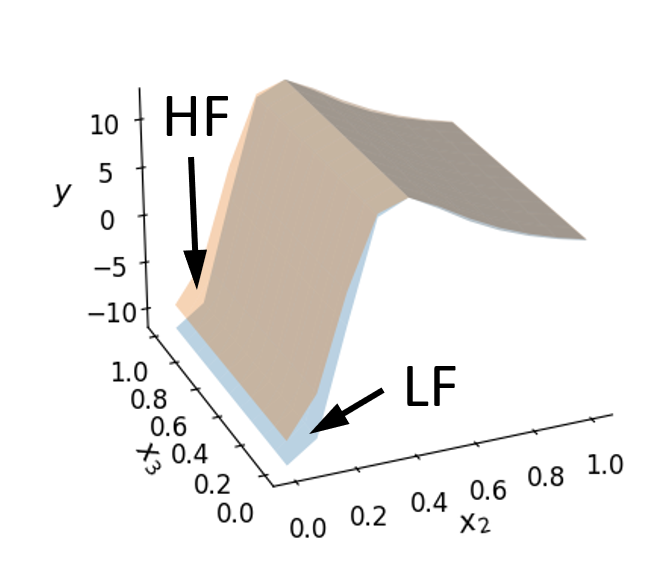}\label{fig:sweeps_c}}
\caption{Comparison of HF and LF CFD models.}
\label{fig:sweeps}
\end{figure}

Lift-to-drag ratio is shown for both the HF and LF CFD models in Fig. \ref{fig:sweeps}. The input variables are scaled to $[-1,1]$. In each plot, the excluded variable is set to the middle of its domain. From the images, the Angle of Attack ($x_2$) is the most influential variable, followed by Mach number ($x_1$), with Altitude ($x_3$) contributing little effect. The models show similar trends, except for Mach number which is linear according to the LF model and quadratic according to the HF model. The captured viscous effects and finer mesh enable the HF model to capture more complexity resulting from the underlying physics.

The problem is formulated as minimizing the negative of the lift-to-drag ratio rather than maximizing the lift-to-drag ratio, following optimization convention. Both the ensemble and GPR models are initialized with $10$ HF samples selected using Latin Hypercube Sampling. Five different optimization runs are completed for each method, using the same $5$ random LHS initializations to ensure a fair comparison. 

\begin{figure}[H]
\centering
\includegraphics[width = 0.6\textwidth]{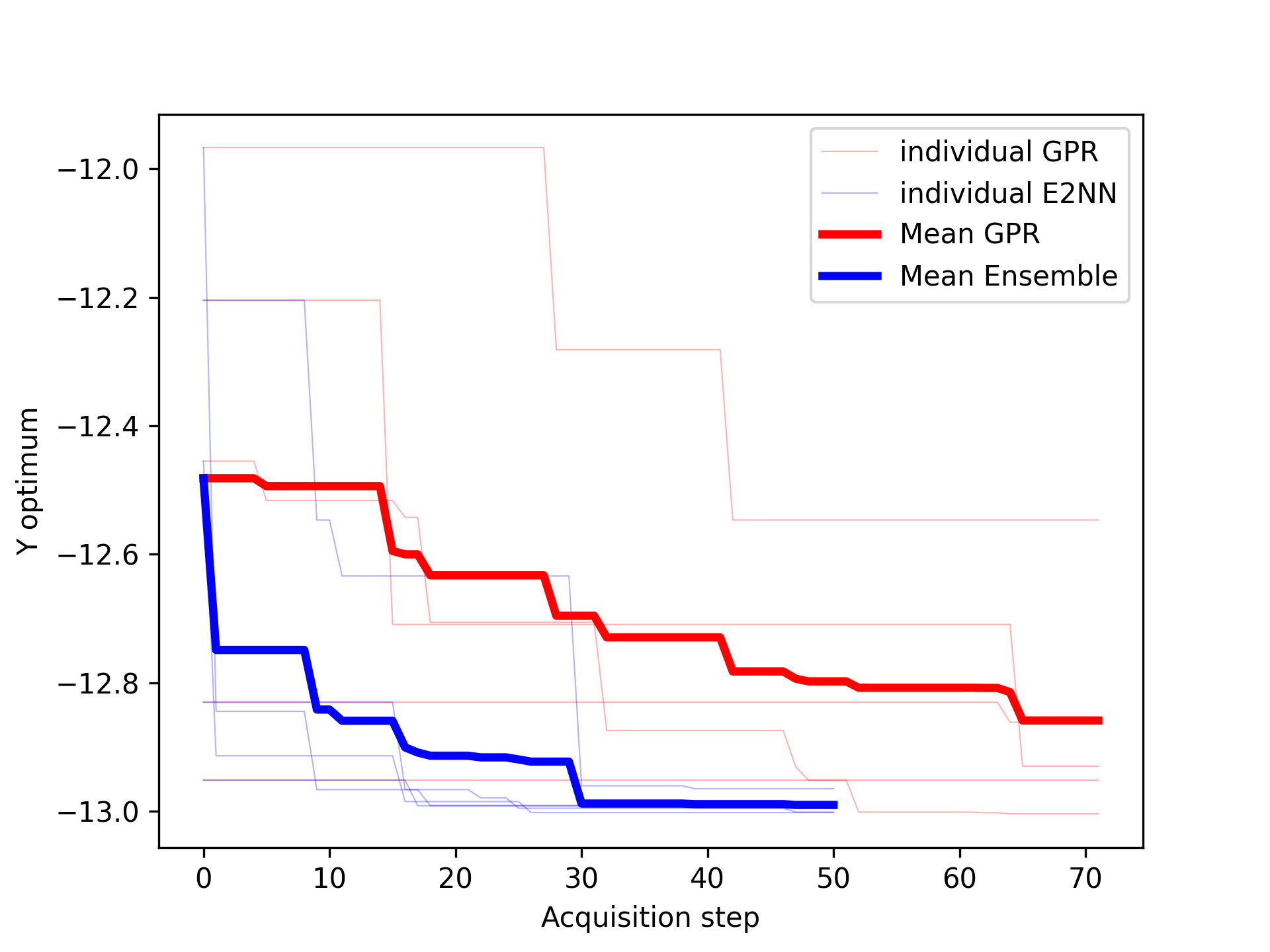}
\caption{Comparison of E2NN and GPR convergence towards optimum $-CL/CD$. Average values across all $5$ runs are shown as thick lines, while the individual run values are shown as thin lines.}
\label{fig:converge}
\end{figure}

The optimization convergence as samples are added is shown in Fig. \ref{fig:converge}. Both models start with the same initial optimum values because they share the same initial design of experiments. However, the E2NN ensemble model improves much more quickly than GPR, and converges to a better optimum. E2NN requires only $11$ HF samples to reach a better optimum on average than GPR reached after $51$ HF samples. In some cases, the GPR model converges prematurely before finding the optimum solution. E2NN much more consistently finds the optimum of $-13$, which corresponds to a lift-to-drag ratio of $13$. 

\section{Conclusions}

In this research, we present a novel framework of adaptive machine learning for engineering design exploration using an ensemble of Emulator Embedded Neural Networks (E2NN) trained as rapid neural networks. This approach allows for the assessment of modeling uncertainty and expedites learning based on the design study's objectives. The proposed framework is successfully demonstrated with multiple analytical examples and an aerospace problem utilizing CFD analysis for a hypersonic vehicle. The performance of the proposed E2NN framework is highlighted when compared to a single-fidelity kriging optimization. E2NN exhibited superior robustness and reduced computational costs in converging to the true optimum. 

The central contribution in this work is a novel technique for approximating epistemic modeling uncertainty using an ensemble of E2NN models. The ensemble predictions are analyzed using Bayesian statistics to produce a t-distribution estimate of the true function response. The uncertainty estimation methodology allows for active learning, which reduces the training data required through efficient goal-oriented design exploration. This is a sequential process of intelligently adding more data based on the ensemble, and then rebuilding the ensemble. The training of the ensemble is drastically accelerated by applying the Rapid Neural Network (RaNN) methodology, enabling individual E2NNs to be trained near-instantaneously. This speedup is enabled by setting some neural network weights to random values and setting others using fast techniques such as linear regression and ridge regression. 

The essential components of the proposed framework are 1) the inclusion of emulators, 2) the ensemble uncertainty estimation, 3) active learning, and 4) the RaNN methodology. These components work together to make the overall methodology feasible and effective. For instance, the active learning would not be possible without the uncertainty estimation. Emulators make the ensemble more robust by stabilizing individual E2NN fits, resulting in fewer defective or outlier fits and better uncertainty estimation. The inclusion of the emulators and the use of adaptive sampling reduces the cost of training data generation more than either method could individually. Finally, the RaNN methodology accelerates training of individual E2NN models by hundreds to thousands of times, preventing the many re-trainings required by the ensemble and adaptive sampling methodologies from introducing significant computational costs. All these techniques combine synergistically to create a robust and efficient framework for engineering design exploration. 

\vspace{1em}
\noindent \fontsize{11}{12} \textbf{Declaration of Competing Interests}
\vspace{0.5em}

The authors declare that they have no known competing financial interests or personal relationships that could have appeared to influence the work reported in this paper.

\vspace{1em}
\noindent \fontsize{11}{12} \textbf{Research Data}
\vspace{0.5em}

No data was used for the research described in this article. The computer codes used for the numerical examples are available at \url{https://github.com/AtticusBeachy/multi-fidelity-nn-ensemble-examples}.

\vspace{1em}
\noindent \fontsize{11}{12} \textbf{Acknowledgement}
\vspace{0.5em}

This research was sponsored by the Air Force Research Laboratory (AFRL) and Strategic Council for Higher Education (SOCHE) under agreement FA8650-19-2-9300. The U.S. Government is authorized to reproduce and distribute reprints for Governmental purposes notwithstanding any copyright notation thereon. The views and conclusions contained herein are those of the authors and should not be interpreted as necessarily representing the official policies or endorsements, either expressed or implied by the Strategic Council for Higher Education and AFRL or the U.S. Government.


\newpage

\bibliographystyle{ieeetr} 
\bibliography{e2nn_ensemble_ArXiv_version}

\begin{thebibliography}{10}

\bibitem{hoerl_ridge_1970}
A.~E. Hoerl and R.~W. Kennard, ``Ridge {Regression}: {Biased} {Estimation} for
  {Nonorthogonal} {Problems},'' {\em Technometrics}, vol.~12, pp.~55--67, Feb.
  1970.
\newblock Publisher: Taylor \& Francis \_eprint:
  https://www.tandfonline.com/doi/pdf/10.1080/00401706.1970.10488634.

\bibitem{tibshirani_regression_1996}
R.~Tibshirani, ``Regression {Shrinkage} and {Selection} via the {Lasso},'' {\em
  Journal of the Royal Statistical Society. Series B (Methodological)},
  vol.~58, no.~1, pp.~267--288, 1996.
\newblock Publisher: [Royal Statistical Society, Wiley].

\bibitem{rumpfkeil_multi-fidelity_2019}
M.~P. Rumpfkeil, D.~E. Bryson, and P.~S. Beran, ``Multi-{Fidelity} {Sparse}
  {Polynomial} {Chaos} {Surrogate} {Models} for {Flutter} {Database}
  {Generation},'' in {\em {AIAA} {Scitech} 2019 {Forum}}, (San Diego,
  California), American Institute of Aeronautics and Astronautics, Jan. 2019.

\bibitem{jones_efficient_1998}
D.~R. Jones, M.~Schonlau, and W.~J. Welch, ``Efficient {Global} {Optimization}
  of {Expensive} {Black}-{Box} {Functions},'' {\em Journal of Global
  Optimization}, vol.~13, no.~4, pp.~455--492, 1998.

\bibitem{moore_value-based_2014}
R.~A. Moore, D.~A. Romero, and C.~J.~J. Paredis, ``Value-{Based} {Global}
  {Optimization},'' {\em Journal of Mechanical Design}, vol.~136, p.~041003,
  Apr. 2014.

\bibitem{bae_nondeterministic_2019}
H.~Bae, D.~L. Clark, and E.~E. Forster, ``Nondeterministic {Kriging} for
  {Engineering} {Design} {Exploration},'' {\em AIAA Journal}, vol.~57,
  pp.~1659--1670, Apr. 2019.

\bibitem{alhazmi_training_2021}
N.~Alhazmi, Y.~Ghazi, M.~N. Aldosari, R.~Tezaur, and C.~Farhat, ``Training a
  {Neural}-{Network}-{Based} {Surrogate} {Model} for {Aerodynamic}
  {Optimization} {Using} a {Gaussian} {Process},'' in {\em {AIAA} {Scitech}
  2021 {Forum}}, (VIRTUAL EVENT), American Institute of Aeronautics and
  Astronautics, Jan. 2021.

\bibitem{peherstorfer_survey_2018}
B.~Peherstorfer, K.~Willcox, and M.~Gunzburger, ``Survey of {Multifidelity}
  {Methods} in {Uncertainty} {Propagation}, {Inference}, and {Optimization},''
  {\em SIAM Review}, vol.~60, pp.~550--591, Jan. 2018.

\bibitem{fischer_bayesian-enhanced_2018}
C.~C. Fischer, R.~V. Grandhi, and P.~S. Beran, ``Bayesian-{Enhanced}
  {Low}-{Fidelity} {Correction} {Approach} to {Multifidelity} {Aerospace}
  {Design},'' {\em AIAA Journal}, vol.~56, pp.~3295--3306, Aug. 2018.

\bibitem{nachar_multi-fidelity_2020}
S.~Nachar, P.-A. Boucard, D.~Néron, and C.~Rey, ``Multi-fidelity bayesian
  optimization using model-order reduction for viscoplastic structures,'' {\em
  Finite Elements in Analysis and Design}, vol.~176, p.~103400, Sept. 2020.

\bibitem{shah_finite_2022}
N.~V. Shah, M.~Girfoglio, P.~Quintela, G.~Rozza, A.~Lengomin, F.~Ballarin, and
  P.~Barral, ``Finite element based {Model} {Order} {Reduction} for
  parametrized one-way coupled steady state linear thermo-mechanical
  problems,'' {\em Finite Elements in Analysis and Design}, vol.~212,
  p.~103837, Dec. 2022.

\bibitem{antil_novel_2021}
H.~Antil, H.~C. Elman, A.~Onwunta, and D.~Verma, ``Novel {Deep} neural networks
  for solving {Bayesian} statistical inverse,'' Feb. 2021.
\newblock Number: arXiv:2102.03974 arXiv:2102.03974 [cs, math].

\bibitem{carlberg_galerkin_2017}
K.~Carlberg, M.~Barone, and H.~Antil, ``Galerkin v. {Least}-{Squares}
  {Petrov}–{Galerkin} {Projection} in {Nonlinear} {Model} {Reduction},'' {\em
  Journal of Computational Physics}, vol.~330, pp.~693--734, Feb. 2017.

\bibitem{boncoraglio_active_2021}
G.~Boncoraglio and C.~Farhat, ``Active {Manifold} and {Model} {Reduction} for
  {Multidisciplinary} {Analysis} and {Optimization},'' in {\em {AIAA} {Scitech}
  2021 {Forum}}, (Virtual Event), American Institute of Aeronautics and
  Astronautics, Jan. 2021.

\bibitem{le_gratiet_recursive_2014}
L.~Le~Gratiet and J.~Garnier, ``Recursive {Co}-{Kriging} {Model} for {Design}
  of {Computer} {Experiments} with {Multiple} {Levels} of {Fidelity},'' {\em
  International Journal for Uncertainty Quantification}, vol.~4, no.~5,
  pp.~365--386, 2014.

\bibitem{ranftl_bayesian_2019}
S.~Ranftl, G.~M. Melito, V.~Badeli, A.~Reinbacher-Köstinger, K.~Ellermann, and
  W.~von~der Linden, ``Bayesian {Uncertainty} {Quantification} with
  {Multi}-{Fidelity} {Data} and {Gaussian} {Processes} for {Impedance}
  {Cardiography} of {Aortic} {Dissection},'' {\em Entropy}, vol.~22, p.~58,
  Dec. 2019.

\bibitem{bae_multifidelity_2020}
H.~Bae, A.~J. Beachy, D.~L. Clark, J.~D. Deaton, and E.~E. Forster,
  ``Multifidelity {Modeling} {Using} {Nondeterministic} {Localized} {Galerkin}
  {Approach},'' {\em AIAA Journal}, vol.~58, pp.~2246--2260, May 2020.

\bibitem{beachy_expected_2020}
A.~J. Beachy, D.~L. Clark, H.~Bae, and E.~E. Forster, ``Expected
  {Effectiveness} {Based} {Adaptive} {Multi}-{Fidelity} {Modeling} for
  {Efficient} {Design} {Optimization},'' in {\em {AIAA} {Scitech} 2020
  {Forum}}, (Orlando, FL), American Institute of Aeronautics and Astronautics,
  Jan. 2020.

\bibitem{cybenko_approximation_1989}
G.~Cybenko, ``Approximation by superpositions of a sigmoidal function,'' {\em
  Mathematics of Control, Signals, and Systems}, vol.~2, pp.~303--314, Dec.
  1989.

\bibitem{hornik_approximation_1991}
K.~Hornik, ``Approximation capabilities of multilayer feedforward networks,''
  {\em Neural Networks}, vol.~4, no.~2, pp.~251--257, 1991.

\bibitem{leshno_multilayer_1993}
M.~Leshno, V.~Y. Lin, A.~Pinkus, and S.~Schocken, ``Multilayer feedforward
  networks with a nonpolynomial activation function can approximate any
  function,'' {\em Neural Networks}, vol.~6, pp.~861--867, Jan. 1993.

\bibitem{peng_multiscale_2021}
G.~C.~Y. Peng, M.~Alber, A.~Buganza~Tepole, W.~R. Cannon, S.~De,
  S.~Dura-Bernal, K.~Garikipati, G.~Karniadakis, W.~W. Lytton, P.~Perdikaris,
  L.~Petzold, and E.~Kuhl, ``Multiscale {Modeling} {Meets} {Machine}
  {Learning}: {What} {Can} {We} {Learn}?,'' {\em Archives of Computational
  Methods in Engineering}, vol.~28, pp.~1017--1037, May 2021.

\bibitem{raissi_physics-informed_2019}
M.~Raissi, P.~Perdikaris, and G.~Karniadakis, ``Physics-informed neural
  networks: {A} deep learning framework for solving forward and inverse
  problems involving nonlinear partial differential equations,'' {\em Journal
  of Computational Physics}, vol.~378, pp.~686--707, Feb. 2019.

\bibitem{meng_composite_2020}
X.~Meng and G.~E. Karniadakis, ``A composite neural network that learns from
  multi-fidelity data: {Application} to function approximation and inverse
  {PDE} problems,'' {\em Journal of Computational Physics}, vol.~401,
  p.~109020, Jan. 2020.

\bibitem{liu_multi-fidelity_2019}
D.~Liu and Y.~Wang, ``Multi-{Fidelity} {Physics}-{Constrained} {Neural}
  {Network} and {Its} {Application} in {Materials} {Modeling},'' {\em Journal
  of Mechanical Design}, vol.~141, p.~121403, Dec. 2019.

\bibitem{aydin_general_2019}
R.~C. Aydin, F.~A. Braeu, and C.~J. Cyron, ``General {Multi}-{Fidelity}
  {Framework} for {Training} {Artificial} {Neural} {Networks} {With}
  {Computational} {Models},'' {\em Frontiers in Materials}, vol.~6, p.~61, Apr.
  2019.

\bibitem{beachy_emulator_2021}
A.~J. Beachy, H.~Bae, I.~Boyd, and R.~V. Grandhi, ``Emulator {Embedded}
  {Neural} {Networks} for {Multi}-{Fidelity} {Conceptual} {Design}
  {Exploration} of {Hypersonic} {Vehicles},'' in {\em {AIAA} {Scitech} 2021
  {Forum}}, (Virtual Event), American Institute of Aeronautics and
  Astronautics, Jan. 2021.

\bibitem{beachy_emulator_2021-1}
A.~Beachy, H.~Bae, I.~Boyd, and R.~Grandhi, ``Emulator embedded neural networks
  for multi-fidelity conceptual design exploration of hypersonic vehicles,''
  {\em Structural and Multidisciplinary Optimization}, July 2021.

\bibitem{bishop_bayesian_1997}
C.~M. Bishop, ``Bayesian {Neural} {Networks},'' {\em Journal of the Brazilian
  Computer Society}, vol.~4, July 1997.

\bibitem{blundell_weight_2015}
C.~Blundell, J.~Cornebise, K.~Kavukcuoglu, and D.~Wierstra, ``Weight
  {Uncertainty} in {Neural} {Networks},'' May 2015.
\newblock Number: arXiv:1505.05424 arXiv:1505.05424 [cs, stat].

\bibitem{ying_zhao_survey_2005}
{Ying Zhao}, {Jun Gao}, and {Xuezhi Yang}, ``A survey of neural network
  ensembles,'' in {\em 2005 {International} {Conference} on {Neural} {Networks}
  and {Brain}}, vol.~1, (Beijing, China), pp.~438--442, IEEE, 2005.

\bibitem{yin_active_2007}
R.~Burbidge, J.~J. Rowland, and R.~D. King, ``Active {Learning} for
  {Regression} {Based} on {Query} by {Committee},'' in {\em Intelligent {Data}
  {Engineering} and {Automated} {Learning} - {IDEAL} 2007} (H.~Yin, P.~Tino,
  E.~Corchado, W.~Byrne, and X.~Yao, eds.), vol.~4881, pp.~209--218, Berlin,
  Heidelberg: Springer Berlin Heidelberg, 2007.
\newblock Series Title: Lecture Notes in Computer Science.

\bibitem{fauser_neural_2015}
S.~Faußer and F.~Schwenker, ``Neural {Network} {Ensembles} in {Reinforcement}
  {Learning},'' {\em Neural Processing Letters}, vol.~41, pp.~55--69, Feb.
  2015.

\bibitem{melville_diverse_2004}
P.~Melville and R.~J. Mooney, ``Diverse ensembles for active learning,'' in
  {\em Twenty-first international conference on {Machine} learning - {ICML}
  '04}, (Banff, Alberta, Canada), p.~74, ACM Press, 2004.

\bibitem{uteva_active_2018}
E.~Uteva, R.~S. Graham, R.~D. Wilkinson, and R.~J. Wheatley, ``Active learning
  in {Gaussian} process interpolation of potential energy surfaces,'' {\em The
  Journal of Chemical Physics}, vol.~149, p.~174114, Nov. 2018.

\bibitem{lin_automatically_2020}
Q.~Lin, Y.~Zhang, B.~Zhao, and B.~Jiang, ``Automatically growing global
  reactive neural network potential energy surfaces: {A} trajectory-free active
  learning strategy,'' {\em The Journal of Chemical Physics}, vol.~152,
  p.~154104, Apr. 2020.

\bibitem{christiano_deep_2017}
P.~Christiano, J.~Leike, T.~B. Brown, M.~Martic, S.~Legg, and D.~Amodei, ``Deep
  reinforcement learning from human preferences,'' {\em arXiv:1706.03741 [cs,
  stat]}, July 2017.
\newblock arXiv: 1706.03741.

\bibitem{schmidt_feedforward_1992}
W.~Schmidt, M.~Kraaijveld, and R.~Duin, ``Feedforward neural networks with
  random weights,'' in {\em Proceedings., 11th {IAPR} {International}
  {Conference} on {Pattern} {Recognition}. {Vol}.{II}. {Conference} {B}:
  {Pattern} {Recognition} {Methodology} and {Systems}}, (The Hague,
  Netherlands), pp.~1--4, IEEE Comput. Soc. Press, 1992.

\bibitem{sivia_data_2012}
D.~S. Sivia and J.~Skilling, {\em Data analysis: a {Bayesian} tutorial}.
\newblock Oxford science publications, Oxford: Oxford Univ. Press, 2. ed.,
  repr~ed., 2012.

\bibitem{tracey_upgrading_2018}
B.~D. Tracey and D.~Wolpert, ``Upgrading from {Gaussian} {Processes} to
  {Student}’s-{T} {Processes},'' in {\em 2018 {AIAA} {Non}-{Deterministic}
  {Approaches} {Conference}}, (Kissimmee, Florida), American Institute of
  Aeronautics and Astronautics, Jan. 2018.

\bibitem{saunders_ridge_1998}
C.~Saunders, A.~Gammerman, and V.~Vovk, ``Ridge {Regression} {Learning}
  {Algorithm} in {Dual} {Variables},'' p.~7, ICML, 1998.

\bibitem{guang-bin_huang_extreme_2004}
{Guang-Bin Huang}, {Qin-Yu Zhu}, and {Chee-Kheong Siew}, ``Extreme learning
  machine: a new learning scheme of feedforward neural networks,'' in {\em 2004
  {IEEE} {International} {Joint} {Conference} on {Neural} {Networks} ({IEEE}
  {Cat}. {No}.{04CH37541})}, vol.~2, (Budapest, Hungary), pp.~985--990, IEEE,
  2004.

\bibitem{huang_extreme_2006}
G.-B. Huang, Q.-Y. Zhu, and C.-K. Siew, ``Extreme learning machine: {Theory}
  and applications,'' {\em Neurocomputing}, vol.~70, pp.~489--501, Dec. 2006.
\newblock Number: 1-3.

\bibitem{huang_universal_2006}
G.-B. Huang, L.~Chen, and C.-K. Siew, ``Universal {Approximation} {Using}
  {Incremental} {Constructive} {Feedforward} {Networks} {With} {Random}
  {Hidden} {Nodes},'' {\em IEEE Transactions on Neural Networks}, vol.~17,
  pp.~879--892, July 2006.

\bibitem{forrester_recent_2009}
A.~I. Forrester and A.~J. Keane, ``Recent advances in surrogate-based
  optimization,'' {\em Progress in Aerospace Sciences}, vol.~45, pp.~50--79,
  Jan. 2009.

\bibitem{rumpfkeil_construction_2017}
M.~P. Rumpfkeil and P.~Beran, ``Construction of {Dynamic} {Multifidelity}
  {Locally} {Optimized} {Surrogate} {Models},'' {\em AIAA Journal}, vol.~55,
  pp.~3169--3179, Sept. 2017.

\bibitem{clark_engineering_2016}
D.~L. Clark, H.-R. Bae, K.~Gobal, and R.~Penmetsa, ``Engineering {Design}
  {Exploration} {Using} {Locally} {Optimized} {Covariance} {Kriging},'' {\em
  AIAA Journal}, vol.~54, pp.~3160--3175, Oct. 2016.

\bibitem{boyd_generic_2020}
I.~Boyd, R.~V. Grandhi, J.~A. Camberos, D.~Sandler, and R.~A. Canfield,
  ``Generic {High}-{Speed} {Vehicle} {Configuration} {Modeling} and
  {Optimization},'' in {\em {AIAA} {AVIATION} 2020 {FORUM}}, (VIRTUAL EVENT),
  American Institute of Aeronautics and Astronautics, June 2020.

\bibitem{anderson_implicit_1994}
W.~Anderson and D.~L. Bonhaus, ``An implicit upwind algorithm for computing
  turbulent flows on unstructured grids,'' {\em Computers \& Fluids}, vol.~23,
  pp.~1--21, Jan. 1994.

\bibitem{murphy_conjugate_nodate}
K.~P. Murphy, ``Conjugate {Bayesian} analysis of the {Gaussian} distribution,''
  {\em https://www.cs.ubc.ca/{\textasciitilde}murphyk/Papers/bayesGauss.pdf}.

\end{thebibliography}

\newpage
\begin{appendices}


\section{Derivation of a Posterior Predictive Distribution from Normally Distributed Data}
\label{section:derivation}

In this section, we derive the posterior predictive distribution given $n$ iid data points $D$ drawn from an underlying normal distribution with unknown mean $\mu$ and variance $\sigma^2$. This proof is an abbreviated version of the proof in \cite{murphy_conjugate_nodate}. First, we need a prior and a likelihood. The prior for the mean and variance is a normal-inverse-chi-squared distribution ($NI\chi^2$) 

\begin{equation}
    \label{eqn:prior}
    p(\mu, \sigma^2) = NI\chi^2(\mu_0,\kappa_0,\nu_0,\sigma_0^2)=N(\mu|\mu_0, \sigma^2/\kappa_0) \cdot \chi^2(\sigma^2|\nu_0,\sigma_0^2)
\end{equation}

Here $\mu_0$ is the prior mean and $\kappa_0$ is the strength of the prior mean, while $\sigma_0^2$ is the prior variance and $\nu_0$ is the strength of the prior variance. The likelihood of the $n$ iid data points is the product of the likelihoods of the individual data points:

\begin{equation}
    \label{eqn:likelihood}
    p(D|\mu, \sigma^2) = \frac{1}{(2\pi)^{n/2}} (\sigma^2)^{-n/2} \exp\left(-\frac{1}{2\sigma^2} \left[n \sum^n_{i=1} (y_i-\bar{y})^2 + n(\bar{y}-\mu)^2 \right] \right)
\end{equation}

By Bayes' rule, the joint posterior distribution for $\mu$ and $\sigma^2$ is 

\begin{equation}
    \label{eqn:posterior}
    p(\mu,\sigma^2|D) \propto  p(D|\mu,\sigma^2)\times p(\mu,\sigma^2)=NI\chi^2(\mu_n, \kappa_n, \nu_n, \sigma^2_n)
\end{equation}

\noindent where $\kappa_n$ and $\nu_n$ can be thought of as the total weight of both the prior and observations.

\begin{equation}
    \label{eqn:kappa_term}
    \kappa_n = \kappa_0+n
\end{equation}

\begin{equation}
    \label{eqn:nu_term}
    \nu_n = \nu_0 + n
\end{equation}

Here $\mu_n$ is a weighted average of the prior and sample means.

\begin{equation}
    \label{eqn:mu_term}
    \mu_n=\frac{\kappa_0\mu_0+n\bar{y}}{\kappa_n}
\end{equation}

Additionally, $\sigma_n^2$ is a weighted average of the prior variance, the sample variance, and the squared difference between the sample mean and the prior mean (which if high implies additional uncertainty in the location of the mean). 

\begin{equation}
    \label{eqn:sigma_term}
    \sigma^2_n = \frac{1}{\nu_n} \left(\nu_0\sigma_0^2+\sum_i(y_i-\bar{y})^2 +\frac{n \kappa_0}{\kappa_0+n}(\mu_0-\bar{y})^2\right)
\end{equation}

Finally, integrating over the mean and variance yields the posterior predictive probability of the response given the sample data

\begin{equation}
    p(y|D)=\int \int p(y|\mu,\sigma^2)p(\mu, \sigma^2|D)~d\mu~d\sigma^2 = \frac{p(y,D)}{p(D)}
\end{equation}

After integration and simplification, this reduces to a t-distribution.

\begin{equation}
    \label{eqn:tdist_general_prior}
    p(y|D)=t_{\nu_n}\left(\frac{1+\kappa_n}{\kappa_n}, \sigma_n^2\right)
\end{equation}

Here a generalized t-distribution with $\nu$ degrees of freedom, a mean of $\mu_t$ and a scale parameter of $\sigma_t$ is written as 

\begin{equation}
    t_\nu(\mu_t, \sigma_t^2)
\end{equation}

The only remaining task is to select appropriate priors for $\mu$ and $\sigma^2$ of the original normal distribution. An uninformative prior can be achieved by setting the mean prior strength $\kappa_0=0$, the variance prior strength $\nu_0=-1$, and the prior variance $\sigma_0^2=0$.

It may seem strange to set the strength of the variance prior $\nu_0$ to $-1$ instead of to $0$ for an uninformative prior. However, because $\kappa_0=0$, estimating the mean requires only one data point, while $\nu_0=-1$ means estimating the variance requires at least two data points. This makes sense: there is no way to even begin to estimate the variance of a distribution from a single data point. With these values selected Eqs. \ref{eqn:kappa_term}-\ref{eqn:sigma_term} become 

\begin{equation}
    \kappa_n=n
\end{equation}

\begin{equation}
    \nu_n=n-1
\end{equation}

\begin{equation}
    \mu_n=\bar{y}
\end{equation}

\begin{equation}
    \sigma_n^2=\frac{1}{n-1}\sum_i(y_i-\bar{y})^2
\end{equation}

Substituting into Eq. \ref{eqn:tdist_general_prior} yields the final pdf for the posterior predictive distribution of a normally distributed random variable $y$ given $n$ iid samples.

\begin{equation}
    \label{eqn:posterior-predictive}
    p(y|D) = t_{n-1}\left(\bar{y}, \frac{1+n}{n}s^2\right)
\end{equation}

\noindent where $\bar{y}$ is the sample mean and $s^2$ is the sample variance. 

\begin{equation}
    s^2 = \frac{1}{n-1}\sum_i(y_i-\bar{y})^2
\end{equation}

For a more detailed example of the derivation of the pdf, see the lecture by Nicholas Zabaras \footnote{Nicholas Zabaras. Lecture 12 - Conjugate Bayesian Analysis Of The Gaussian (Part A). \url{https://www.youtube.com/watch?v=CTKIa8V4-6g}. Accessed Apr. 28, 2022.}. Another derivation is performed in \cite{sivia_data_2012} (section 3.3.1, pg. 54) using different reasoning, although that example only goes as far as deriving the marginal posterior distribution of $\mu$, and does not derive the posterior predictive distribution of $y$. In that derivation, uninformative priors are used for $\mu$ and $\sigma$. The mean $\mu$ is uniformly distributed on the interval $(-\infty,\infty)$ and the log of the standard deviation $\log(\sigma)$ is also uniformly distributed on the interval $(-\infty,\infty)$. These prior distributions are used because they are the distributions of maximum entropy. The observed data are then used to perform Bayesian updating of the joint distribution of $\mu$ and $\sigma$ (they are not independent). 

%
%
%
%
%

\end{appendices}

\end{document}